\documentclass[letterpaper]{article} 
\usepackage{aaai25}  
\usepackage{times}  
\usepackage{helvet}  
\usepackage{courier}  
\usepackage[hyphens]{url}  
\usepackage{graphicx} 
\usepackage{natbib}  
\usepackage{caption} 
\usepackage{algorithm}
\usepackage{algorithmic}
 
\usepackage{newfloat}
\usepackage{listings}
\DeclareCaptionStyle{ruled}{labelfont=normalfont,labelsep=colon,strut=off} 
\floatstyle{ruled}
\newfloat{listing}{tb}{lst}{}
\floatname{listing}{Listing}
\usepackage{tabularx} 
\usepackage{booktabs} 
\usepackage{multirow} 

\usepackage[symbol]{footmisc}
\DefineFNsymbols{myfnsymbols}{{\ensuremath{*}}{\ensuremath{\circ}}}
\setfnsymbol{myfnsymbols}

\title{REVECA: Adaptive Planning and Trajectory-based Validation in Cooperative Language Agents using Information Relevance and Relative Proximity}

\author{
    SeungWon Seo\equalcontrib\textsuperscript{\rm 1},
    SeongRae Noh\equalcontrib\textsuperscript{\rm 1},
    Junhyeok Lee\textsuperscript{\rm 1},
    SooBin Lim\textsuperscript{\rm 1}, \\
    Won Hee Lee\textsuperscript{\rm 1}, 
    HyeongYeop Kang\thanks{Corresponding author.}\textsuperscript{\rm 2}
}
\affiliations{
    \textsuperscript{\rm 1}Kyung Hee University, Yongin, Republic of Korea\\
    \textsuperscript{\rm 2}Korea University, Seongbuk-gu, Republic of Korea\\
    \{ssw03270, rhosunr99, bluehyena123, dnpcs, 	whlee\}@khu.ac.kr, 
    \{siamiz\_hkang\}@korea.ac.kr
}

\begin{document}

\maketitle
\begin{abstract}
We address the challenge of multi-agent cooperation, where agents achieve a common goal by cooperating with decentralized agents under complex partial observations. Existing cooperative agent systems often struggle with efficiently processing continuously accumulating information, managing globally suboptimal planning due to lack of consideration of collaborators, and addressing false planning caused by environmental changes introduced by other collaborators. To overcome these challenges, we propose the \textbf{RE}levance, Proximity, and \textbf{V}alidation-\textbf{E}nhanced \textbf{C}ooperative Language \textbf{A}gent (REVECA), a novel cognitive architecture powered by GPT-4o-mini. REVECA enables efficient memory management, optimal planning, and cost-effective prevention of false planning by leveraging Relevance Estimation, Adaptive Planning, and Trajectory-based Validation. Extensive experimental results demonstrate REVECA's superiority over existing methods across various benchmarks, while a user study reveals its potential for achieving trustworthy human-AI cooperation. 
\end{abstract}
%

\section{Introduction}
Digital agents collaborating with humans are crucial in games, educational platforms, and virtual universes. 
Commonly referred to as Non-Player Characters, these agents play a crucial role in enhancing user immersion in commercial applications, where they assist users in achieving desired objectives. However, these agents operate on pre-scripted behaviors, limiting their adaptability to complex scenarios and rich conversations with humans.
Inspired by the reasoning and communication capabilities of Large Language Models (LLMs)~\cite{Li2023CAMELCA}, we aim to create agents that use LLMs for more effective cooperation in complex environments and improved communication with humans, surpassing traditional methods relying on static, learnable models or reinforcement learning (RL).

This paper introduces REVECA, a \textbf{RE}levance, Proximity, and \textbf{V}alidation-\textbf{E}nhanced \textbf{C}ooperative Language \textbf{A}gent, an LLM-based agent framework addressing decentralized control, costly communication, complex tasks, partially observable environments, and noisy settings.
Similar to prior work~\cite{zhang2023building}, we mainly focus on VirtualHome~\cite{Puig2018VirtualHomeSH}, the multi-objective household tasks using well-constructed virtual settings. Specifically, we focus on two decentralized agents cooperating on a multi-objective, long-horizon household task under complex partial observations. 
Additionally, continuous communication incurs time costs, and irrelevant dummy objects add noise, complicating the environment. 

The primary advancements of REVECA over previous works are threefold. 
Firstly, REVECA significantly reduces both the computational complexity and memory demands of the planning by prioritizing information based on its relevance to the task objectives. This allows the agent to focus on the most pertinent data, thereby maintaining robust performance even in noisy environments typical of real-world scenarios. Information relevance is assessed at the point of acquisition, utilizing the reasoning capabilities of LLMs.
Previous research~\cite{zhang2023building,Li2023SemanticallyAT} stored all scene information in memory, but this led to performance issues due to the fixed context window of LLMs. Some studies ~\cite{Zhang2023ProAgentBP,Zhang2024COMBOCW} used queues to retain recent $K$ pieces of information, but this led to suboptimal planning due to limited historical data.

Secondly, REVECA enhances plan optimality by incorporating the relative proximity between collaborators and task objectives. This relative proximity is assessed by inferring potential interaction between collaborators and task objectives, utilizing LLMs and agent's observation data.
Previous research on distributed cooperation within partially observed environments~\cite{zhang2023building, Zhang2023ProAgentBP} has shown that individual agents may be unable to determine whether their optimal plan aligns with the group's best outputs, often leading to suboptimal collective outcomes. 

Lastly, REVECA mitigates the occurrence of false plans by implementing a cost-effective plan validation process.
In partially observable environments with multiple task objectives, a collaborator's task completion may not be immediately updated in the agent's memory, leading to the creation of redundant plans.
To address this, REVECA estimates the likelihood that a collaborator has already completed a given task by inferring potential interaction trajectories. This inference is based on LLMs and historical observation data to predict the collaborators' trajectories, spanning from the last retrieval of the collaborator's information to the current time. 
Traditional methods~\cite{Li2023TheoryOM} rely on constant communication for updates, which is costly and prohibitive when cooperating with humans.

To demonstrate REVECA's contributions, this paper presents the results of comparative analysis, ablation studies, and user studies conducted in three multi-room simulation environments: Communicative Watch-And-Help (C-WAH)~\cite{zhang2023building}, ThreeDWorld Multi-Agent Transport (TDW-MAT)~\cite{zhang2023building}, and Noisy-C-WAH. Noisy-C-WAH, a variant of C-WAH with dummy obstacles that are interactive but unrelated to the task objectives, was used to evaluate REVECA's performance in noisy conditions. 
To further assess the generalization capability of our architecture in a fully observable environment without any modification, the Overcooked-AI~\cite{carroll2019utility} environment is also included in the experiments. 
Results demonstrate that REVECA outperforms recent approaches in terms of success rates, efficiency, and robustness.

\section{Related Work}
\subsection{Cooperative Agents with RL}
Research on cooperative agents has a long history ~\cite{stone2000multiagent,gronauer2022multi}. 
The traditional cooperative agent has been studied across various directions, mainly leveraging RL techniques to enable cooperation with diverse collaborators and adapt to dynamic environments~\cite{misra2018mapping,Amato2019ModelingAP,jaderberg2019human,strouse2021collaborating,yu2022surprising,zhao2023maximum,li2023cooperative,li2024tackling,zhong2024heterogeneous}.
These approaches have facilitated the development of agents that can autonomously learn to cooperate by maximizing cumulative rewards through trial and error in various simulated settings. Notable studies have explored aspects such as mapping state spaces to actions effectively~\cite{misra2018mapping} and enhancing the robustness of learning algorithms in multi-agent systems~\cite{Amato2019ModelingAP}.

To evaluate these approaches, some other researchers have aimed to develop platforms that can test the performance of cooperative agents~\cite{lowe2017multi,savva2019habitat,xiang2020sapien,puig2020watch,padmakumar2022teach,li2023behavior,zhou2024hazard}.
These provide standardized environments to benchmark the performance and generalization capabilities of the agents. 

\subsection{Planning and Decision Making with LLMs}
Despite the various studies aimed at developing cooperative agents, a major limitation of previous work has been the lack of natural language-based communication between agents~\cite{Das2018TarMACTM,carroll2019utility,jaderberg2019human,puig2020watch}.
Effective language-based communication is crucial for enhancing collaboration, particularly in complex multi-agent environments and when collaborating with humans~\cite{lazaridou2016multi}.

Recently, the advanced reasoning and natural language processing capabilities of LLMs have significantly enhanced agents' decision-making~\cite{li2022pre,wang2023voyager,huang2023grounded,yuan2023distilling} and planning abilities~\cite{Huang2022InnerME,huang2022language,Li2023TheoryOM,wang2024describe}.

The integration of LLMs has also significantly improved the development of cooperative LLM-based embodied agents~\cite{zhang2023building,Zhang2023ProAgentBP,Li2023SemanticallyAT,Zhang2024COMBOCW}. 
These agents utilize LLMs to understand the environment, plan tasks, and facilitate communication with both human users and other agents.
However, existing studies encounter suboptimal performance due to inherent challenges associated with LLMs, such as performance degradation when processing large volumes of input data~\cite{Levy2024SameTM}, and inadequate reasoning abilities in handling complex reasoning tasks~\cite{Ullman2023LargeLM}.
Furthermore, these studies struggle with the issue of false planning in decentralized and dynamic multi-agent environments. One potential solution involves employing plan evaluation strategies~\cite{madaan2024self, Shinn2023ReflexionLA}; however, they have primarily been explored within static, single-agent environments. Another solution is to implement constant communication between collaborators~\cite{Li2023TheoryOM}; however, this incurs substantial communication overhead, making it particularly impractical in scenarios involving human collaborators~\cite{zhang2023building}. 

To address the limitations of previous works, we introduce REVECA, an LLM-based cooperative embodied agent framework. REVECA facilitates efficient memory management, optimal planning, and cost-effective prevention of false planning by leveraging information relevance, relative proximity, and plan validation.

\section{Problem Definition}
The problem setting of our work is an extension of the decentralized partially observable Markov decision process (DEC-POMDP). Following previous conventions~\cite{Bernstein2000TheCO, Spaan2006DecentralizedPU, zhang2023building, Zhang2024COMBOCW}, our problem is defined as follows. In a state space $S$, $N$ agents collaborate to achieve a \textit{common goal} $G = \{g_1, \ldots, g_v\}$, which consists of \textit{v} sub-goals.
$M$ and $A$ are the memory set and action set of the agent. 
The memory structure is defined as $M = M_o \cup M_c$, where $M_o$ is \textit{observation memory}, containing object information $I_o$ prioritized based on its relevance to the $G$, with relevance scores $R = \{Strong, Medium, Low, None\}$. \textit{Strong}, assigned by the agent at observation time, indicates the highest priority.
$M_c$ is \textit{collaborator memory}, containing collaborator information $I_c$. It consists of information about collaborators, including both historical data extracted from the conversation logs and directly observed information in a decentralized, partially observable environment.
It encompasses a sequence of communication messages $\sigma_i$ from the respective collaborator, annotated with the corresponding simulation step $i$.
The Action set $A= A_c \cup A_l$ comprises $A_c$, which consists of actions related to the transmission of a message $\sigma$ to a collaborator, and $A_l$, which consists of pre-defined actions essential for task execution, as specified in the \textit{low-level action skill book}.
For simulation steps $ i = \{1, \ldots, H\}$, the state transition function $T$ governs the transition from state $s_i$ to state $s_{i+1}$ based on the action $a_i$, denoted as $T(s_{i+1},s_i,a)=p(s_{i+1}|s_i,a_i)$. 
The probability of an agent taking action $a_i$ according to the plan $\pi_i$ in state $s_i$ is given by $p(a_i|s_i)=p(a_i|\pi_i, s_i)p(\pi_i|s_i, M_i)$. 
The simulation runs for a maximum of $H$ steps and will terminate under any of the following conditions: the completion of all sub-goals, reaching the step limit $H$, or the depletion of viable actions for all agents.

\begin{figure*}[t!]
    \centering 
    \includegraphics[width=\linewidth]{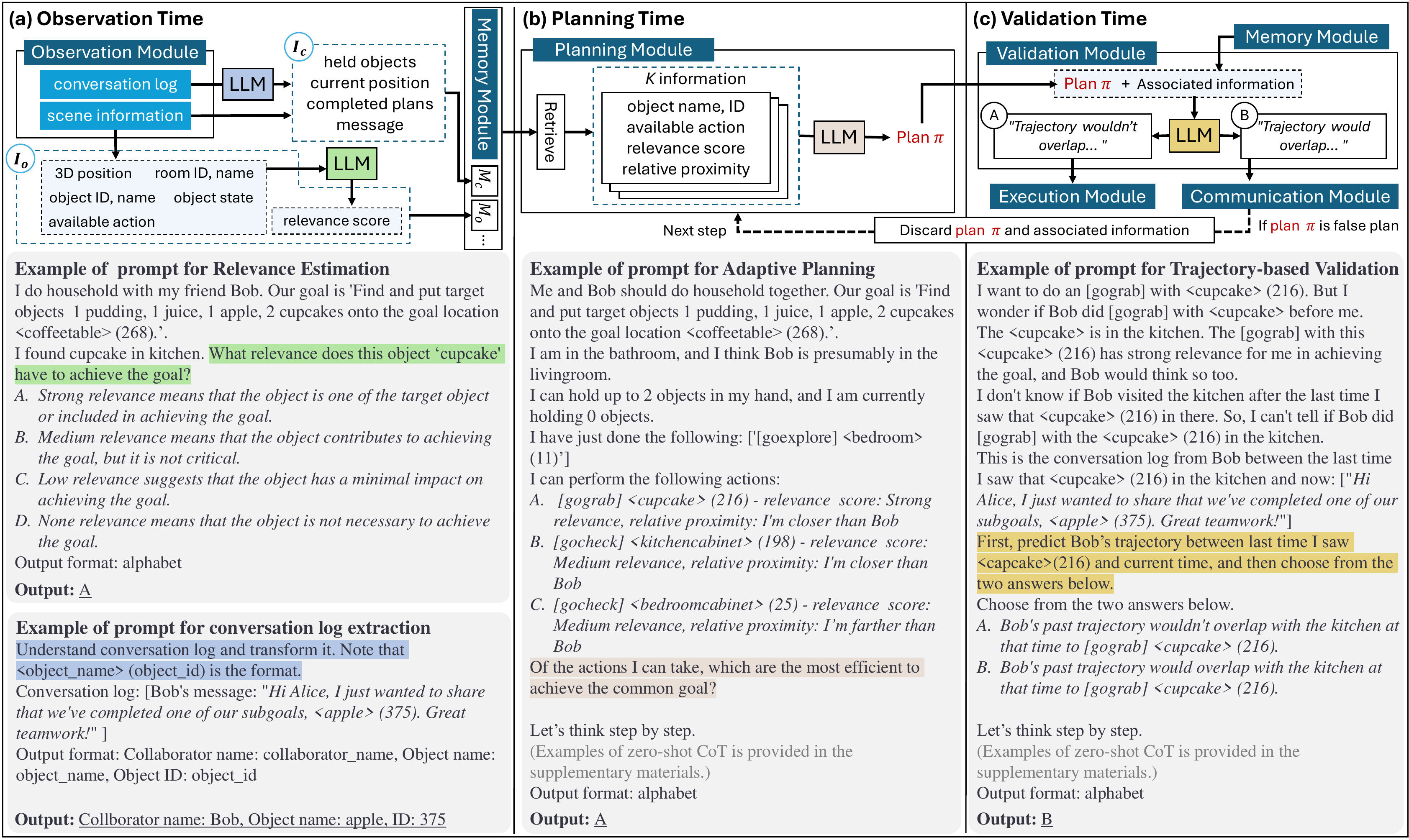}
    \caption{The REVECA process workflow ensures efficient memory management, optimal planning, and cost-effective prevention of false planning through three phases: (a) Observation Time, (b) Planning Time, and (c) Validation Time.}
\label{fig:main_figure}
\end{figure*}

\section{REVECA framework}
Our framework comprises six modules: 1) Communication Module, 2) Observation Module, 3) Memory Module, 4) Planning Module, 5) Validation Module, and 6) Execution Module. These modules incorporate three key processes: Relevance Estimation, Adaptive Planning, and Trajectory-based Validation.

\textbf{Communication Module} facilitates information sharing between agents through natural language, leveraging the advanced capabilities of recent LLMs~\cite{Bubeck2023SparksOA}.

\textbf{Observation Module} is responsible for collecting and categorizing environmental data into four levels of relevance, based on what the agent can observe from its current location. Note that the partial observation environment restricts the agent's observational scope.

\textbf{Memory Module} comprises four components: \textit{common goal}, \textit{observation memory}, \textit{collaborator memory}, and a \textit{low-level action skill book}. This module is responsible for storing, updating, and managing data critical to the agent's decision-making processes.

\textbf{Planning Module} retrieves $K$ information data from the $M_o$ and adaptively generates the plan $\pi$, using relevance scores and relative proximity. 

\textbf{Validation Module} estimates the likelihood that a collaborator has already completed a given task goal by predicting the collaborator's trajectories. If the plan generated by the Planning Module involves a task goal that has a high probability of being completed by a collaborator, the plan is identified as a false plan, prompting its reformulation.

\textbf{Execution Module} executes validated plans using the pre-defined \textit{low-level action skill book}, which are implemented in Python. The skill book is provided by the benchmark framework.

These modules are invoked throughout REVECA's iterative workflow phases: Observation Time, Planning Time, and Validation Time. 
Further details on the modular design of REVECA are provided comprehensively in the supplementary materials, and a demonstration is showcased in the accompanying video.

\subsection{Communication for information sharing}

The Communication Module, which facilitates information sharing through natural language, is invoked in four cases. First, at the initiation of the simulation, all agents exchange their initial positions and information regarding surrounding objects. Second, when an agent requires another agent's task history for validation purposes. Third, when an agent needs to provide its task history in response to a validation request. Finally, when a sub-goal is completed, the achievement is announced to all other agents.

\subsection{Observation Time: Relevance Estimation}
During the Observation Time, the Observation Module acquires raw scene information and refines it into $I_o$. $I_o$ encompasses details about objects such as their 3D positions, object IDs and names, room IDs and names, available actions, and object states. In a multi-room environment, the agent's observations are restricted to objects and collaborators within its current room; objects within closed containers (e.g., cabinets or boxes) remain unobserved until accessed. $I_o$ is assigned a $R$, evaluated by LLMs based on the $G$. This relevance-based prioritization avoids the reference of all memory entries and simplifies the LLM-based planning process by focusing on the most pertinent information.

$I_c$ is obtained either through direct observation of the collaborator or via natural language communication with other agents. After the communication session terminates, $I_c$ is refined by integrating relevant details extracted from the conversation log using LLMs with information obtained through direct observation. $I_c$ includes the collaborator's held objects, current position, message, and the history of completed plans.

Both $I_o$ and $I_c$ are stored within the Memory Module as $M_o$ and $M_c$, respectively. 
An example of how the relevance score is determined and how data is extracted from a communication message is depicted in Figure \ref{fig:main_figure}(a).



\subsection{Planning Time: Adaptive Planning}
The Planning Module begins by retrieving $K$ pieces of $I_o$ based on their relevance score and the agent's relative proximity to the associated objects. 
The relative proximity is calculated as the current distance between the agent, the object position stored in $I_o$, and the most recent positions of the collaborators stored in $I_c$.
First, all instances of $I_o$ stored in $M_o$ are sorted in descending order according to their relevance score $R$. 
For information entries with identical relevance scores, prioritization is further refined by calculating the relative proximity $P$. 
Although proximity is naturally a continuous measure, we found that converting numerical distance into corresponding natural language descriptions (e.g. I'm closer than Bob, I'm farther than Bob) significantly enhances LLMs performance. LLMs then utilize zero-shot chain-of-thought prompting (CoT)~\cite{Kojima2022LargeLM} to generate the plan $\pi$ by including the retrieved $K$ pieces of $I_o$, relevance score, and relative proximity in the input prompt, as illustrated in Figure \ref{fig:main_figure}(b). To facilitate this process, the agent's current information including the held object, current position, and completed plan history is incorporated to prompt as additional context.

This approach implicitly guides the Planning Module in generating a globally optimal plan among collaborators based on relevance scores and relative proximity.

\subsection{Validation Time: Trajectory-based Validation}
Even a well-constructed plan can become invalid due to environmental changes caused by collaborators during the interval between observation and planning. 
In partially observable environments, detecting such changes poses a significant challenge for the agent.

A straightforward method to resolve this issue is to revisit the object's location or query all collaborators about their interactions. However, this can lead to inefficient path planning and incur substantial communication costs, which is particularly impractical when collaborating with humans.

To address this, REVECA's Validation Module incorporates Trajectory-based Validation, which estimates the validity of a plan using both $M_o$ and $M_c$. To validate a plan $\pi$ generated by the Planning Module, the agent predicts each collaborator's past trajectory $\tau_{i}$, where $ 1 \leq i \leq N-1$, covering the period from the acquisition time of information used for planning $\alpha$ to the Planning Time $\beta$, where $1 \leq \alpha \leq \beta \leq H$. 
To construct $\tau_i$, all relevant information about collaborator $i$ is retrieved from $M_c$, along with the relevance scores and $I_o$ from $M_o$, specifically focusing on the information stored between simulation steps $\alpha$ and $\beta$.
Given the discontinuity in the associated $I_o$ used in the plan, LLMs’ reasoning capabilities are leveraged to infer the missing information, thereby constructing trajectory $\tau_{i}$.

Based on the plan validity check, if it is determined that no collaborator is likely to have interacted with the object, the agent assumes the plan is valid and proceeds with execution. Otherwise, the agent first sends a message via the Communication Module to the collaborator with the highest interaction probability to confirm the prediction. If the collaborator confirms the interaction, it indicates that the current plan $\pi$ is a false plan. Consequently, the agent discards both the $\pi$ and associated $I_o$ used in its formulation and then proceeds with a new planning process during the planning time of the next step. 
If the collaborator denies the interaction, the agent discards the corresponding $\tau$ and queries the next collaborator with the second-highest interaction probability, repeating this process until no potential candidates remain. If all collaborators deny the interaction, indicating current plan $\pi$ is not a false plan, the agent deems the plan $\pi$ valid and proceeds with its execution.
It is important to note that our environment assumes all agents share a $G$, possess equal capabilities, and act cooperatively. Therefore, we do not consider scenarios where an agent, despite being fully capable, chooses not to interact with an object necessary for achieving the $G$.
An example of Trajectory-based Validation is depicted in Figure \ref{fig:main_figure}(c).

\subsection{Executing Navigation and Contextual Actions}
Once the plan is finalized, the Execution Module retrieves $I_o$ from the $M_o$ to identify the target location. 
For efficient pathfinding, the A-star search algorithm is employed to navigate the agent toward the object.  
Upon approaching the object, the agent retrieves an available action from the \textit{low-level action skill book} to execute the planned interaction.

\section{Experiment}
We conducted experiments performing multi-objective household tasks using three types of indoor multi-room simulation environments: C-WAH~\cite{zhang2023building}, TDW-MAT~\cite{zhang2023building}, and Noisy-C-WAH—all of which are partially observable. Additionally, we used the cooperative game simulation, Overcooked-AI~\cite{carroll2019utility}, which is fully observable, to further evaluate the framework's performance under conditions where complete information is available.

In C-WAH and Noisy-C-WAH, we evaluate the agent performance using Simulation Steps (SS) and Travel Distance (TD) to measure the time cost to achieve the $G$ and the average distance traveled, respectively. 
We conducted 10 episodes, each with 3 to 5 sub-goals, across two environments, with $H$ set to 250 steps.

In TDW-MAT, performance is evaluated based on the success rate of transporting items, including the overall success rate (TOTAL), and specific success rates for objects categorized as Food (FOOD) and Stuff (STUFF). Each category includes 10 target objects, and $H$ is set to 3000 steps.

In Overcooked-AI, we evaluate the agent performance based on the reward, where a reward of 20 points is obtained each time the two agents successfully complete and serve a dish. We conducted 5 different layouts and $H$ is set to 400 steps. Layout images and detailed descriptions of all experiment settings are provided in the supplementary materials.

\subsection{REVECA and Baselines}

In the comparative experiments conducted in partial observation, our REVECA was evaluated against three baselines: the MCTS-based Hierarchical Planner (MHP), the Rule-based Hierarchical Planner (RHP), and the Cooperative Embodied Language Agent (CoELA). We compared REVECA with MHP and CoELA in C-WAH, with RHP and CoELA in TDW-MAT, and with CoELA in Noisy-C-WAH. 

In the comparative experiments conducted in full observation Overcooked-AI, REVECA is evaluated against six baselines: self-play (SP)~\cite{tesauro1994td, carroll2019utility}, Population Based Training (PBT)~\cite{jaderberg2017population}, Fictitious Co-Play (FCP)~\cite{strouse2021collaborating}, Maximum Entropy Population-based training (MEP)~\cite{zhao2023maximum}, Cooperative Open-ended LEarning (COLE)~\cite{li2023cooperative, li2024tackling}, and the ProAgent~\cite{Zhang2023ProAgentBP}. 

We evaluate the robustness of cooperation between different methods by generating 49 pairs of combinations using our method, REVECA, and the six baselines. The order of the first and second players was reversed for each combination to account for varying starting positions, resulting in the full set of pairs. 

To further evaluate the robustness across different versions of LLMs, we conducted experiments using \textit{gpt-4o-mini-2024-07-18} (4o-mini), \textit{gpt-3.5-turbo-0125}, and \textit{Meta-Llama-3.1-8B-Instruct} (Llama 3.1) on the C-WAH and Noisy-C-WAH environments. For all other environments, only 4o-mini was utilized. 
Additionally, we incorporated GPT-4-driven CoELA performance from the CoELA manuscript~\cite{zhang2023building} to facilitate a more comprehensive analysis of the capacities of different LLMs. 
Detailed descriptions of all baseline models and LLMs versions are provided in the supplementary materials to ensure reproducibility. 

\subsection{Comparative Results: In Partial Observation}

\begin{table}[t!]
    \centering
    \begin{tabularx}{\columnwidth}{lccccccc}
    \toprule
        Method & LLMs & & & & SS $\downarrow$ & & TD (m) $\downarrow$ \\
     \midrule
        MHP & \multicolumn{1}{c}{X} & & & & 69.40 & & 58.96 \\
        CoELA & GPT-3.5 & & & & 71.90 & & 61.29 \\
        CoELA & 4o-mini & & & & 65.50 & & 53.64 \\
        CoELA & GPT-4 & & & & 57.00 & & / \\
    \midrule
        REVECA & Llama 3.1 & & & & 56.00 & & 47.13 \\
        REVECA & GPT-3.5 & & & & 48.90 & & 40.34 \\
        \textbf{REVECA} & \textbf{4o-mini} & & & & \textbf{44.20} & & \textbf{38.44} \\
     \bottomrule
    \end{tabularx}
    \caption{Comparative experimental results in C-WAH environment. The best result is highlighted in bold.}
    \label{tab:C-WAH_Result}
\end{table}

\begin{table}[t!]
    \centering
    \begin{tabularx}{\columnwidth}{lcccc}
    \toprule
         Method & LLMs & TOTAL$\uparrow$ & FOOD$\uparrow$ & STUFF$\uparrow$ \\
     \midrule
        RHP & \multicolumn{1}{c}{X} & 0.79 & 0.83 & 0.76 \\
        CoELA & 4o-mini & 0.53 & 0.51 & 0.55 \\
        CoELA & GPT-4 & 0.71 & 0.82 & 0.61 \\
    \midrule
        \textbf{REVECA} & \textbf{4o-mini} & \textbf{0.87} & \textbf{0.87} & \textbf{0.87} \\
     \bottomrule
    \end{tabularx}
    \caption{Comparative experimental results in TDW-MAT environment. The best result is highlighted in bold.}
    \label{tab:TDW_MAT_Result}
\end{table}

\begin{figure}[t!]
    \centering 
    \includegraphics[width=\linewidth]{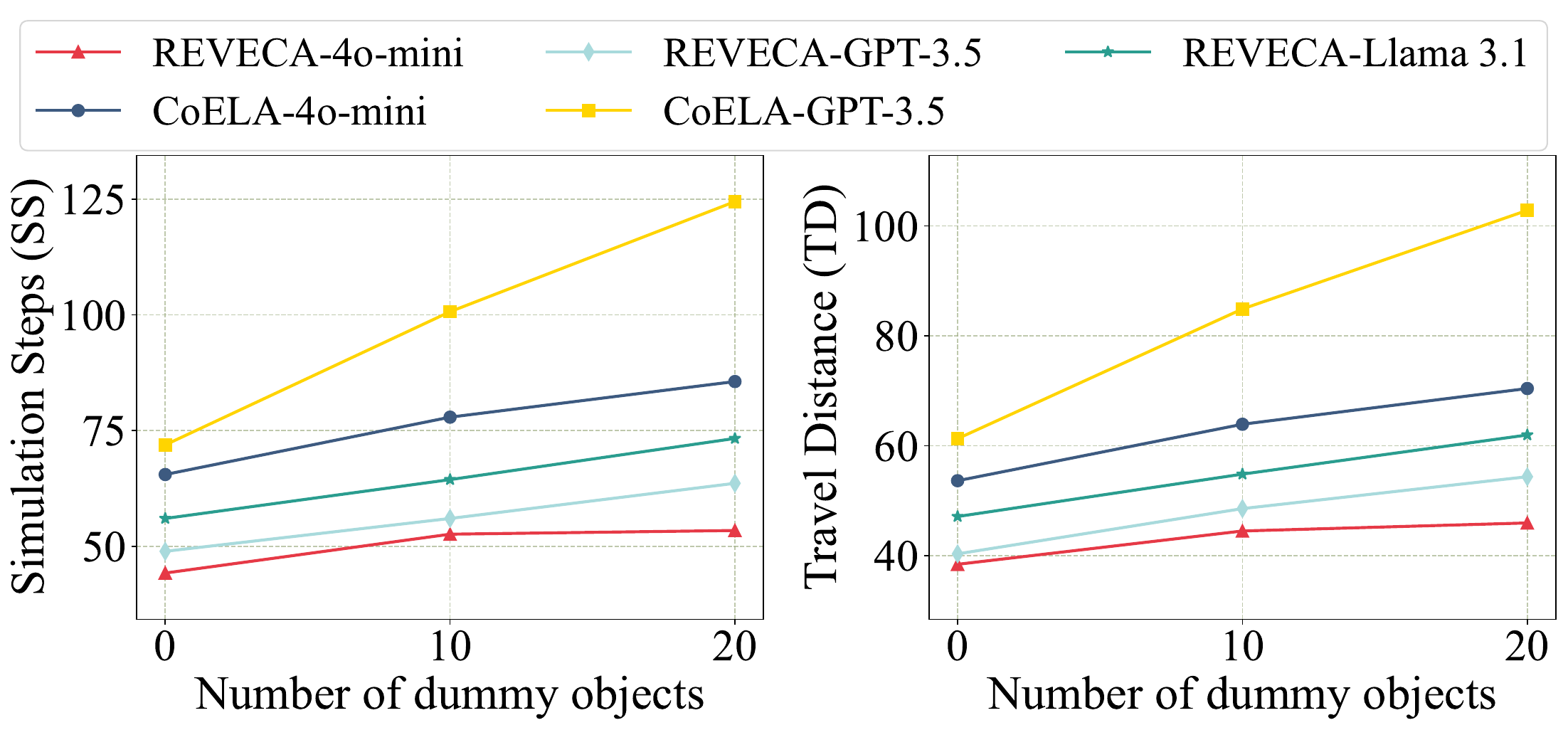}
    \caption{Comparative results in the Noisy-C-WAH environment with varying dummy objects.}
    \label{fig:Noisy-C-WAH_Results}
\end{figure}

Table \ref{tab:C-WAH_Result} presents that REVECA outperforms other baseline methods in the C-WAH environment. Notably, the REVECA driven by various LLMs require fewer SS and less TD to complete tasks compared to GPT-4, 4o-mini, GPT-3.5-driven CoELA, and MHP, thereby demonstrating superior efficiency. 
Interestingly, the performance of GPT-3.5-driven CoELA falls below that of MHP, and GPT-4-driven CoELA underperforms compared to the Llama 3.1-driven REVECA. This demonstrates CoELA's significant dependence on GPT-4's advanced reasoning capabilities.

Table \ref{tab:TDW_MAT_Result} presents results from the TDW-MAT environment, where 4o-mini-driven REVECA outperforms all baselines, including GPT-4-driven CoELA, across all metrics (TOTAL, FOOD, and STUFF).

In the Noisy-C-WAH environment, the experiment included 10 or 20 additional dummy objects. As shown in Figure \ref{fig:Noisy-C-WAH_Results}, REVECA models driven by various LLMs outperform CoELA driven by GPT-3.5 and 4o-mini across all metrics. CoELA's strategy of storing all acquired information in text form leads to a significant decline in reasoning performance, a limitation that becomes even more pronounced when weaker LLMs are used for reasoning. As the number of dummy objects increases, the benefits of utilizing relevance scores become increasingly evident.

\begin{table*}[t!]
    \centering
    \small
    \begin{tabularx}{\textwidth}{lcccccccccc}
    \toprule
         \multirow{2}{*}{Method} & \multicolumn{2}{c}{Cramped Room} & \multicolumn{2}{c}{Asymmetric Advantage} & \multicolumn{2}{c}{Coordination Ring} & 
         \multicolumn{2}{c}{Forced Coordination} & 
         \multicolumn{2}{c}{Counter Circuit} \\
         & Agent1 $\uparrow$ & Agent2 $\uparrow$ & Agent1 $\uparrow$ & Agent2 $\uparrow$ & Agent1 $\uparrow$ & Agent2 $\uparrow$ & Agent1 $\uparrow$ & Agent2 $\uparrow$ & Agent1 $\uparrow$ & Agent2 $\uparrow$ \\
     \midrule
        SP & 165.71 & 174.29 & 174.29 & 197.14 & 122.86 & 111.43 & 34.29 & 48.57 & 77.14 & 68.57 \\
        PBT & \textbf{182.86} & 185.71 & 197.14 & 185.71 & 142.86 & 142.86 & 62.86 & \textbf{91.43} & 74.29 & 57.14 \\
        FCP & 171.43 & \textbf{188.57} & 177.14 & 177.14 & 122.86 & 142.86 & 45.71 & 40.00 & 57.14 & 40.00\\
        MEP & 180.00 & \textbf{188.57} & 157.14 & 197.14 & 174.29 & 154.29 & 25.71 & 42.86 & 68.57 & 77.14 \\
        COLE & 177.14 & 151.43 & 205.71 & 188.57 & \textbf{177.14} & 174.29 & 37.14 & 54.29 & 85.71 & 108.57 \\
        ProAgent & 165.71 & 140.00 & 260.00 & \textbf{254.29} & 162.86 & 174.29 & 85.71 & 40.00 & \textbf{125.71} & \textbf{131.43} \\
     \midrule
        \textbf{REVECA} & 157.14 & 171.43 & \textbf{262.86} & 234.29 & 174.29 & \textbf{177.14} & \textbf{88.57} & 62.86 & 120.00 & 125.71 \\
     \bottomrule
    \end{tabularx}
    \caption{Comparative study results between REVECA and baselines in Overcooked-AI. The best result is highlighted in bold.}
    \label{tab:Overcooked_Result}
\end{table*}

\begin{table*}[t!]
    \centering
    \small
    \begin{tabularx}{\textwidth}{lcccccccccc}
    \toprule
         \multirow{2}{*}{Method} & \multicolumn{2}{c}{Cramped Room} & \multicolumn{2}{c}{Asymmetric Advantage} & \multicolumn{2}{c}{Coordination Ring} & 
         \multicolumn{2}{c}{Forced Coordination} & 
         \multicolumn{2}{c}{Counter Circuit} \\
         & Agent1 $\uparrow$ & Agent2 $\uparrow$ & Agent1 $\uparrow$ & Agent2 $\uparrow$ & Agent1 $\uparrow$ & Agent2 $\uparrow$ & Agent1 $\uparrow$ & Agent2 $\uparrow$ & Agent1 $\uparrow$ & Agent2 $\uparrow$ \\
     \midrule
        SP & 100.00 & 80.00 & 120.00 & 60.00 & 48.00 & 40.00 & 28.00 & 12.00 & 36.00 & 32.00 \\
        PBT & 96.00 & 92.00 & 124.00 & 64.00 & 68.00 & 64.00 & \textbf{52.00} & 4.00 & 40.00 & 28.00 \\
        FCP & 148.00 & 148.00 & 160.00 & 44.00 & 112.00 & 100.00 & 28.00 & 32.00 & 12.00 & 20.00 \\
        MEP & \textbf{164.00} & 152.00 & 160.00 & 64.00 & \textbf{144.00} & 104.00 & 44.00 & 32.00 & 36.00 & 44.00 \\
        COLE & 148.00 & 136.00 & 164.00 & 204.00 & 96.00 & 92.00 & 48.00 & 48.00 & 90.00 & 84.00 \\
        ProAgent & 160.00 & \textbf{156.00} & 212.00 & \textbf{240.00} & 140.00 & 120.00 & 24.00 & 88.00 & 108.00 & \textbf{124.00} \\
     \midrule
        \textbf{REVECA} & 132.00 & 140.00 & \textbf{216.00} & 208.00 & 132.00 & \textbf{128.00} & \textbf{52.00} & \textbf{96.00} & \textbf{112.00} & 112.00 \\
     \bottomrule
    \end{tabularx}
    \caption{Comparative study results with BC Models in Overcooked-AI. The best result is highlighted in bold.}
    \label{tab:Overcooked_BC_Result}
\end{table*}

\subsection{Comparative Results: In Full Observation}

Table \ref{tab:Overcooked_Result} presents Overcooked-AI results, where REVECA, driven by 4o-mini, exhibits high cooperative performance in a fully observable environment, achieving comparable to the state-of-the-art ProAgent, which also utilizes 4o-mini. 
Notably, our method is not specialized for fully observable environments, demonstrating REVECA's versatility. This finding suggests that REVECA is broadly applicable, not only to household tasks but also to cooperative multi-agent game environments governed by specialized game rules, without requiring any modifications.

As presented in previous research~\cite{Zhang2023ProAgentBP, carroll2019utility}, we further conducted comparative experiments in Overcooked-AI by training behavior cloning (BC) models using human data to simulate human users. In this experiment, we tested all pairwise combinations of five BC models and other methods, including REVECA.

As shown in Table \ref{tab:Overcooked_BC_Result}, REVECA demonstrates superior performance in collaboration with various BC models trained on human data, achieving the highest scores more frequently than ProAgent. Specifically, in the Forced Coordination layout, REVECA consistently achieved high performance regardless of the starting positions. This demonstrates REVECA's robustness in specialized scenarios where each worker is restricted to their designated workspace and cannot substitute for others. This highlights its adaptability and effectiveness in such unique environments.

\subsection{Ablation Study Results}

\begin{table}[t!]
    \centering
    \small
    \begin{tabularx}{\columnwidth}{lccccc}
    \toprule
         \multirow{2}{*}{Method} & & \multicolumn{2}{c}{C-WAH} & \multicolumn{2}{c}{Noisy-C-WAH} \\
         & & SS $\downarrow$ & TD (m) $\downarrow$ & SS $\downarrow$ & TD (m) $\downarrow$ \\
     \midrule
        full observation & & 45.50 & 37.26 & 52.10 & 42.50 \\
     \midrule
        w/o CoT & & 48.90 & 43.47 & 63.90 & 55.17 \\
        w/o proximity & & 71.60 & 64.02 & 74.40 & 66.18 \\
        w/o other info & & 48.80 & 42.33 & 65.00 & 57.39 \\
    \midrule
        w/o relevance & & 46.20 & 38.84 & 82.10 & 68.52 \\
        w/o validation & & 45.80 & 40.02 & 54.60 & 47.27 \\
    \midrule
        $K$ = 1 & & 45.20 & 39.17 & 57.20 & 49.01 \\
        $K$ = 2 & & 45.60 & 39.91 & 63.60 & 53.69 \\
        $K$ = 4 & & 47.40 & 40.96 & 53.50 & 45.60 \\
        $K$ = Inf & & 53.70 & 46.45 & 88.90 & 77.51 \\
    \midrule
        $R$ = 3 & & 47.20 & 40.20 & 68.30 & 57.94 \\
        $R$ = 5 & & 54.90 & 46.75 & 76.90 & 64.86 \\
    \midrule
        \textbf{REVECA} & & \textbf{44.20} & \textbf{38.44} & \textbf{53.40} & \textbf{45.96} \\
     \bottomrule
    \end{tabularx}
    \caption{Ablation study results in the environments of C-WAH and Nosiy-C-WAH augmented by 20 dummy objects. The best result is highlighted in bold, with the exception of full observation.}
    \label{tab:REVECA_Ablation_Result}
\end{table}

To demonstrate the significance of each component in our framework, we conducted an ablation study within C-WAH and Noisy-C-WAH environments, the latter augmented with 20 dummy objects.
The results are presented in Table \ref{tab:REVECA_Ablation_Result}.

Initially, we tested REVECA in a fully observable setting by enforcing communication before executing any action (full observation). In this setting, each agent is forced to broadcast its perceived information to all other agents before executing any action. While this consumes SS for communication, the agents always maintain up-to-date information and therefore do not generate false plans. One might question whether forced communication is the most convenient approach, but the subsequent user study reveals that this approach is not well-suited for collaboration with humans.

We also tested REVECA under three modified conditions: without using CoT (w/o CoT), without considering relative proximity (w/o proximity), and without using other agents' information (w/o other info). These factors are critical to the reasoning process in REVECA. The experimental results showed a decline in both SS and TD, with the most pronounced reduction occurring when relative proximity was excluded. This finding indicates that the absence of relative proximity impairs the generation of globally efficient paths for the embodied agent, as it fails to consider the collaborator's movements.

Next, we evaluated REVECA without using relevance scores, instead employing a distance-based greedy search approach (w/o relevance). While this approach only slightly underperformed compared to REVECA in C-WAH, likely due to the relatively small number of items, it resulted in the lowest performance across all scores in Noisy-C-WAH, except in the $K$ = Inf setting. 
Additionally, we conducted an experiment by removing the Validation Module (w/o validation), which led to increased SS and TD scores, indicating a performance decline. This decline is attributable to the agent's inability to prevent false plans, thereby hindering the creation of efficient collaborative trajectories.

Lastly, we varied the number of plans $K$ and relevance levels $R$ considered by the LLM planner. The default parameters of $K$ = 3 and $R$ = 4 yielded the optimal performance. In this context, $K$ = 1 corresponds to retrieving only a single piece of information, while $K$ = Inf represents bypassing the retrieval process entirely, thereby imposing the maximum computational load on the LLM planner, as it must directly reference the entire memory.

\begin{figure}[t!]
    \centering 
    \includegraphics[width=\linewidth]{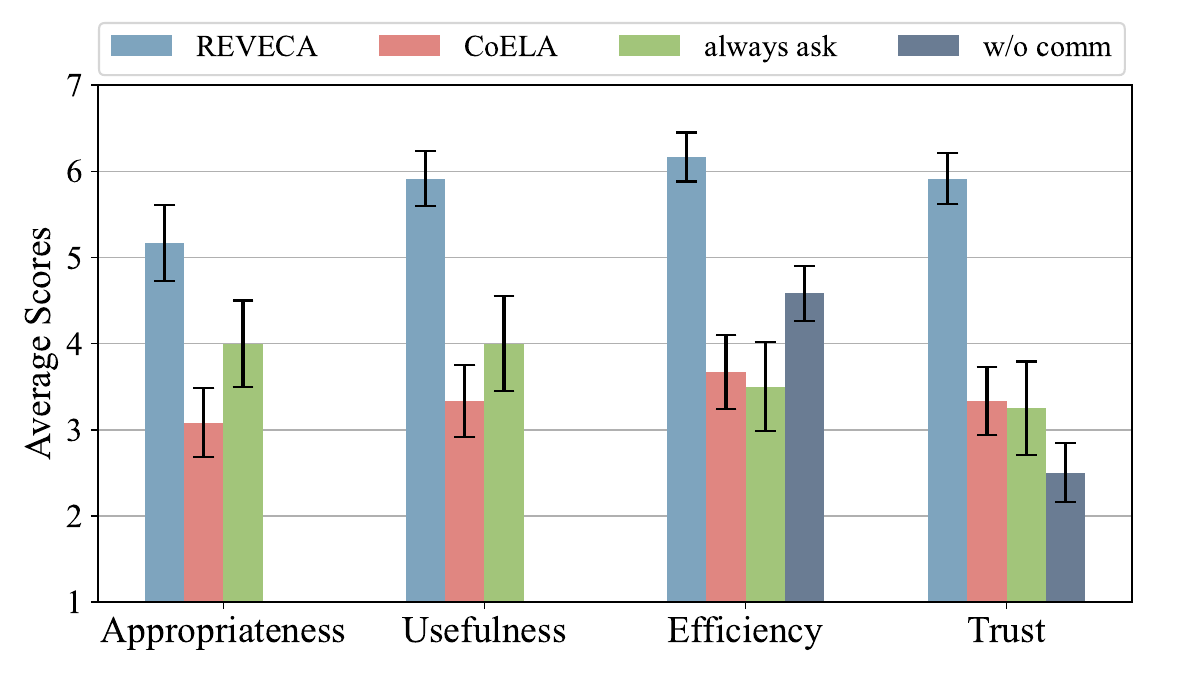}
    \caption{User study results in C-WAH environment. The mean scores and associated standard errors for responses to four research questions.}
    \label{fig:user_study_results}
\end{figure}

\subsection{User Study Results}
We conducted a user study to evaluate REVECA's ability to collaborate seamlessly with humans to achieve $G$.
Twelve participants (nine men and three women) with an average age of 23.67 years were recruited. 
The experiment took place in the C-WAH environment using four methods: REVECA, CoELA, REVECA with an ``always ask before action" approach (always ask), and REVECA without communication (w/o comm).

Participants shared the same observation and action space as the agents, interacting with the environment by selecting actions from a predefined list. 
Each participant completed five sub-goals with each method.
After completing each method, they answered a 7-point Likert scale (1: strongly disagree, 7: strongly agree) questionnaire that addressed four key research questions: 1) Did the agent respond appropriately to your intentions? (Appropriateness), 2) Was the interaction with the agent helpful in achieving the goal? (Usefulness), 3) Did the agent's performance help achieve the goal quickly? (Efficiency), and 4) Did you feel a sense of trust with the agent? (Trust)
The ``w/o comm" method excluded questions on Appropriateness and Usefulness, due to the lack of interaction.
Following the questionnaire, participants were interviewed to gather qualitative feedback on each method.

As shown in Figure \ref{fig:user_study_results}, REVECA scored highest across all four questions, demonstrating its superior performance in human-agent collaboration. 
Participants noted that CoELA frequently produced messages focused on status reports and planning, rather than directly addressing their questions, which led to lower scores in Appropriateness and Usefulness. 
In the ``always ask" condition, participants found the agent's repetitive questions, which were often of low relevance to their current actions, to be disruptive and demotivating
Regarding trust, participants noted that the lack of communication in the ``w/o comm" method made it difficult to understand the agent's actions and situation, thereby hindering trust-based collaboration. Further analysis of the user study is provided in the supplementary materials.

\section{Conclusion}

In this paper, we introduced REVECA, an LLM-driven cognitive architecture designed for multi-objective household tasks, enabling efficient cooperation between decentralized agents under complex, partially observable environments. By leveraging Relevance Estimation, Adaptive Planning, and Trajectory-based Validation, REVECA enhances agent cooperation in dynamic settings while minimizing communication costs, making it well-suited for human collaboration and effectively managing irrelevant dummy objects. Furthermore, we demonstrate REVECA's generalization capacity in a fully observable game environment. However, REVECA has several limitations. Its effectiveness in open-world outdoor settings with constantly changing remains to be validated. Using \textit{low-level action skill book} could be enhanced by integrating recent advancements in character animation generation technologies. Addressing these limitations could make future versions of REVECA even more robust and applicable across a broader range of multi-agent environments and tasks.

\section*{Acknowledgements}
This work was supported by ICT Creative Consilience Program through the Institute of Information \& Communications Technology Planning \& Evaluation(IITP) grant funded by the Korea government(MSIT)(IITP-2024-RS-2020-II201819), Institute of Information \& communications Technology Planning \& Evaluation (IITP) grant funded by the Korea government(MSIT) (No.RS-2020-II200861), the Ministry of Science and ICT (MSIT), Korea, through the ITRC (Information Technology Research Center) support program (IITP-2024-RS-2024-00438239), the Global AI Frontier Lab project (No. RS-2024-00509257), and the Artificial Intelligence Convergence Innovation Human Resources Development program (No. RS-2022-00155911, Kyung Hee University), all supervised by the Institute for Information \& Communications Technology Planning \& Evaluation (IITP).

\bibliography{aaai25}
\newpage

\twocolumn[{%
\renewcommand\twocolumn[1][]{#1}%
\maketitle
\section{\textbf{\LARGE Appendices}}
\section{}
\begin{center}
    \centering
    \captionsetup{type=figure}
    \includegraphics[width=\textwidth]{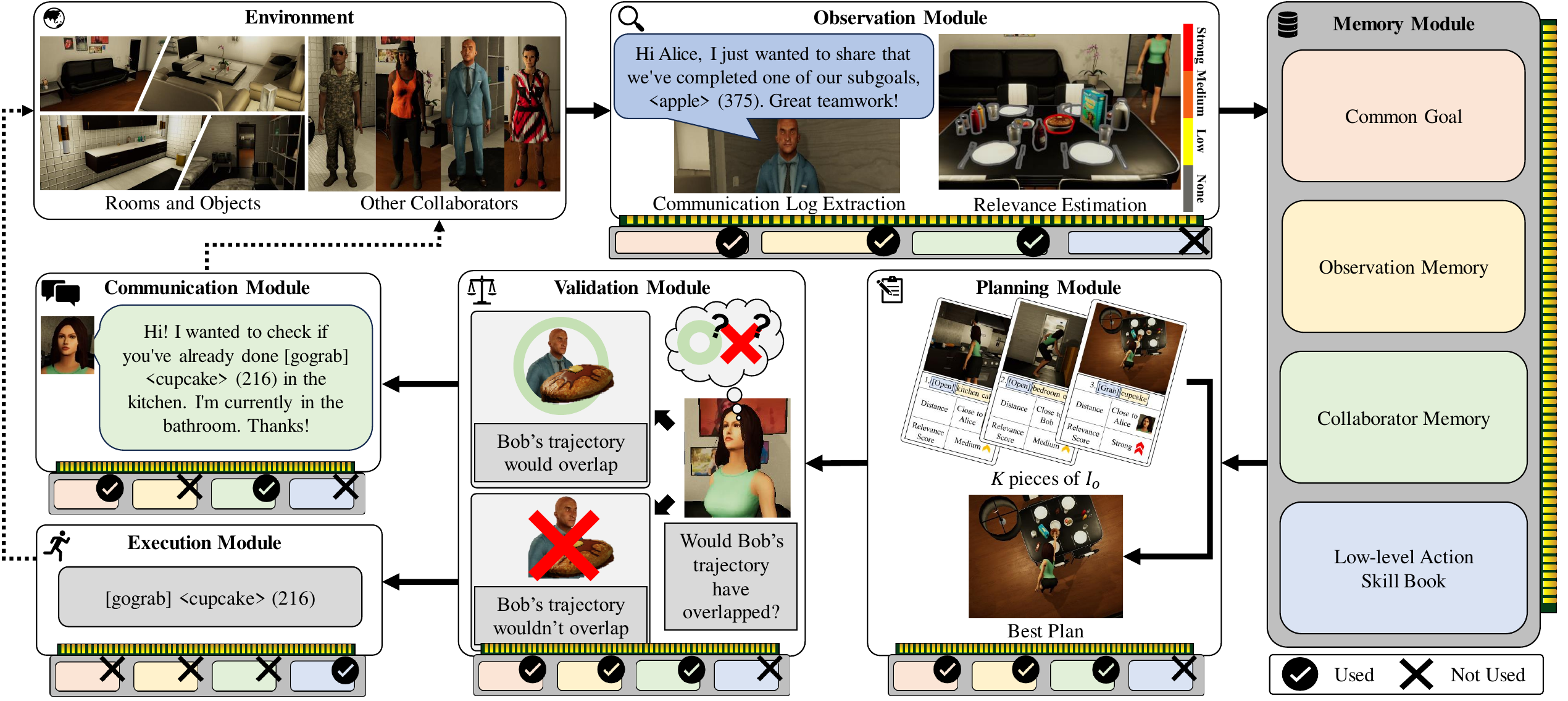}
    \captionof{figure}{An example scenario demonstrating REVECA’s comprehensive operational flow, highlighting the interaction between various modules, collaborators, and the environment to achieve a \textit{common goal}.}
    \label{fig:framework}
\end{center}%
}]

\section{Index}
Our supplementary material is organized as follows:
\begin{itemize}
    \item Additional details and examples of information stored in the Memory Module.
    \item Layout images and detailed descriptions of experiment settings.
    \item Detailed descriptions of baseline models and LLMs.
    \item Additional analysis of the user study and inference times.
    \item Additional detail about key processes of REVECA.
    \item Discussions regarding complex environments.
    \item Full-version of prompts within the REVECA framework.
    \item Scenarios highlighting the importance of each module.
\end{itemize}
The operational flow of REVECA illustrated through an example scenario is shown in Figure \ref{fig:framework}.

\section{Example of Memories in Memory Modules}
In this section, we provide detailed examples of memories in each Memory Module: \textit{common goal}, \textit{observation memory}, \textit{collaborator memory}, and \textit{low-level action skill book}.  

\begin{listing}[t!]%
\caption{Example of \textit{common goal}}%
\label{lst:common_goal}%
\begin{lstlisting}[language=python]
# A simple example of the common goal.
common_goal = "Find and put target objects 1 pudding, 1 juice, 1 apple, 2 cupcakes onto the goal location <coffeetable> (268)."
\end{lstlisting}
\end{listing}

\begin{listing}[t!]%
\caption{Example of \textit{observation memory}}%
\label{lst:observation_memory}%
\begin{lstlisting}[language=python]
# A simple example of an object information list with one entry.
object_information_list = [{
    "object_id": 21, 
    "object_name": 'apple', 
    "position": [13.22, 1.20, 5.41], 
    "available_action": 'gograb', 
    "room_name": 'livingroom', 
    "room_id": 198, 
    "states": [15, "GRABBABLE"] 
}]
\end{lstlisting}
\end{listing}


\begin{listing}[t!]%
\caption{Example of \textit{collaborator memory}}%
\label{lst:collaborator_memory}%
\begin{lstlisting}[language=python]
# A simple example of a collaborator information list with one entry.
collaborator_information_list = [{
    "held_object_ids": [380, 385], 
    "position": [8.78, 1.25, -4.70],
    "conversation_log": 
        [{
            step: 22, 
            message: "Hi Alice, I just wanted to share that we've completed one of our subgoals, <cupcake> (368). Great progress!"
        }]
    "completed_plans": 
        ["[gograb] <cupcake> (368)", "[goput]<coffeetable> (268)"]
}]
\end{lstlisting}
\end{listing}

\begin{listing}[t!]%
\caption{Example of \textit{low-level action skill book}}%
\label{lst:low_level_skill_book}%
\begin{lstlisting}[language=python]
# A simple example of a low-level action skill book.
def goexplore(self):
    if current_room == target_room:
    # Move towards a specific room based on the plan.
    
def gocheck(self):
    if 'OPEN' in container['states']:
    # Check the status of a container and attempt to interact with it.
    
def gograb(self):
    if target in reachable_objects:
    # Attempt to grab an object, ensuring availability and conditions.

def goput(self):
    if len(grabbed_objects) > 0:
    # Place the grabbed object in the specified location or container.

\end{lstlisting}
\end{listing}

\subsection{Common Goal}
The \textit{common goal} $G$ is stored within the Memory Module, described through natural language. An example of this is shown in Listing~\ref{lst:common_goal}.

\subsection{Observation Memory}
Object information $I_o$ is stored within the Memory Module as \textit{observation memory}
$M_o$, consists of details about objects, such as their 3D positions, object IDs and names, room IDs and names, available actions, and object states. An example of this is shown in Listing~\ref{lst:observation_memory}. 



\subsection{Collaborator Memory}
Collaborator information $I_c$ is stored within the Memory Module as \textit{collaborator memory}
$M_c$, which encompasses the collaborator's held objects, current position, messages, and the history of completed plans. An example of this is shown in Listing~\ref{lst:collaborator_memory}. Notably, since the collaborator is not always under observation, the agent cannot keep all information about the collaborator up to date. 
Therefore, we extract the room location of the collaborator from the conversation log and use this information to update the collaborator’s position to the coordinates of the room's center. 
For example, if REVECA extracts information from the conversation log and determines that the collaborator is currently in the kitchen, the collaborator’s current coordinates are updated to the center coordinates of the kitchen. This method enables REVECA to roughly estimate the collaborator’s location and calculate the relative proximity.

\subsection{Low-level Action Skill Book}
The \textit{low-level action skill book} is stored within the Memory Module, which is implemented in Python. An example of this is shown in Listing~\ref{lst:low_level_skill_book}.

\section{Experiment Settings}
In our experiments, the agent's observations are conducted using the APIs provided by each environment, consistent with previous studies~\cite{zhang2023building, Zhang2023ProAgentBP}. We conduct experiments in partially observable environments such as C-WAH, Noisy-C-WAH, and TDW-MAT, as well as in the fully observable environment Overcooked-AI. Below, we provide detailed descriptions of each environment.

\subsection{C-WAH and Noisy-C-WAH Environments}
C-WAH~\cite{zhang2023building}, an extended version of the Watch-And-Help Challenge~\cite{puig2020watch}, is built on the realistic multi-agent simulation platform, VirtualHome~\cite{Puig2018VirtualHomeSH}. This environment includes five household tasks, such as setting the table and doing the dishes, which form the \textit{common goal}, as listed in Table~\ref{tab:c-wah_goal}. Our experiments consist of 10 episodes, each with five different household tasks across two test environments. In C-WAH, agents can acquire or provide information through communication with other agents while executing instructions. 
When an agent enters a specific room, it can observe all objects not inside a container such as fridges or microwaves. To observe objects inside containers, the agent must perform an additional action of opening them. 
Agents are limited to using up to 500 characters per frame to mimic real-world communication costs. 
The horizon $H$ is set to 250 simulation steps, and each task includes 3 to 5 subgoals (or objects). Failing to achieve the \textit{common goal} within the 250 steps results in an episode failure. 

\begin{figure}[t!]
    \centering 
    \includegraphics[width=\linewidth]{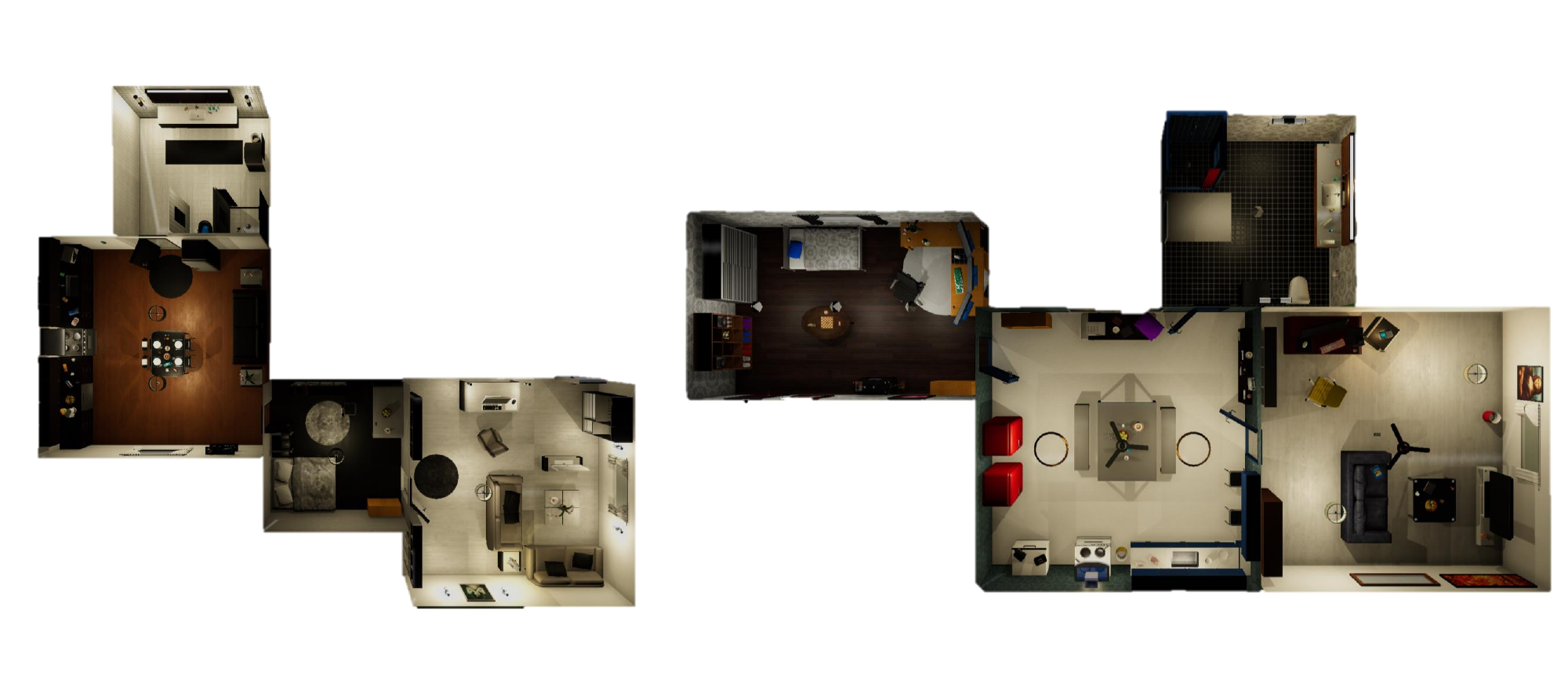}
    \caption{The example layouts of the C-WAH and Noisy-C-WAH environments.}
\label{fig:c-wah_env}
\end{figure}

Noisy-C-WAH is a customized version of C-WAH, augmented with an additional 10 or 20 dummy objects per episode to demonstrate the robustness of our framework in noisy environments. These dummy objects increase the complexity of the environment, challenging the agents' observation, planning, and communication processes. The layout is shown in Figure~\ref{fig:c-wah_env} and the detailed action space is listed in Table~\ref{tab:c-wah_action}.

\begin{table}[t!]
    \centering
    \begin{tabularx}{\columnwidth}{>{\raggedright\arraybackslash}p{0.375\columnwidth} >{\raggedright\arraybackslash}p{0.55\columnwidth}}
    \toprule
        Goals & Description \\
     \midrule
        Prepare afternoon tea & put [\textit{cupcake, pudding, apple, juice, wine}] on \textit{coffeetable} \\
    \addlinespace
        Wash dishes & put [\textit{plate, fork}] inside \textit{dishwasher} \\
    \addlinespace
        Prepare a meal & put [\textit{coffeepot, cupcake, pancake, poundcake, pudding, apple, juice, wine}] on \textit{dinnertable} \\
    \addlinespace
        Put groceries & put [\textit{cupcake, pancake, poundcake, pudding, apple, juice, wine}] inside \textit{fridge} \\
    \addlinespace
        Set up a dinner table & put [\textit{plate, fork}] on \textit{dinnertable} \\
     \bottomrule
    \end{tabularx}
    \caption{The goal specifications of the C-WAH and Noisy-C-WAH environments.}
    \label{tab:c-wah_goal}
\end{table}

\begin{table}[t!]
    \centering
    \begin{tabularx}{\columnwidth}{>{\raggedright\arraybackslash}p{0.25\columnwidth} >{\raggedright\arraybackslash}p{0.65\columnwidth}}
    \toprule
        Action & Description \\
     \midrule
        Walk towards & move to an object in the same room with the agents or a room \\
    \addlinespace
        Turn left & turn left by 30 degrees \\
    \addlinespace
        Turn right & turn right by 30 degrees \\
    \addlinespace
        Grasp & grasp an object \\
    \addlinespace
        Open & open a closed container \\
    \addlinespace
        Close & close an open container \\
    \addlinespace
        Put & put the held objects into an open container or onto a surface  \\
    \addlinespace
        Send message & send a message to other agents \\
     \bottomrule
    \end{tabularx}
    \caption{The action space of the C-WAH and Noisy-C-WAH environments.}
    \label{tab:c-wah_action}
\end{table}

\begin{table}[t!]
    \centering
    \begin{tabularx}{\columnwidth}{>{\raggedright\arraybackslash}p{0.075\columnwidth} >{\raggedright\arraybackslash}p{0.175\columnwidth} >{\raggedright\arraybackslash}p{0.65\columnwidth}}
    \toprule
        Task & Type & Object Name \\
     \midrule
        \multirow{3}{*}{Food} & Containers & [\textit{bowl, plate, tea\_tray}] \\
         & \multirow{2}{*}{Objects} & [\textit{bread, burger, loaf\_bread, apple, banana, orange}] \\
     \midrule
        \multirow{4}{*}{Stuff} & \multirow{2}{*}{Containers} & [\textit{plastic\_basket, wicker\_basket, wood\_basket}] \\
         & \multirow{2}{*}{Objects} & [\textit{iPhone, pen, key, iPod, lighter, purse, calculator, pencil\_bucket, mouse}] \\
     \bottomrule
    \end{tabularx}
    \caption{The target objects and containers of the TDW-MAT environments.}
    \label{tab:tdw_obj}
\end{table}

\begin{figure}[t!]
    \centering 
    \includegraphics[width=\linewidth]{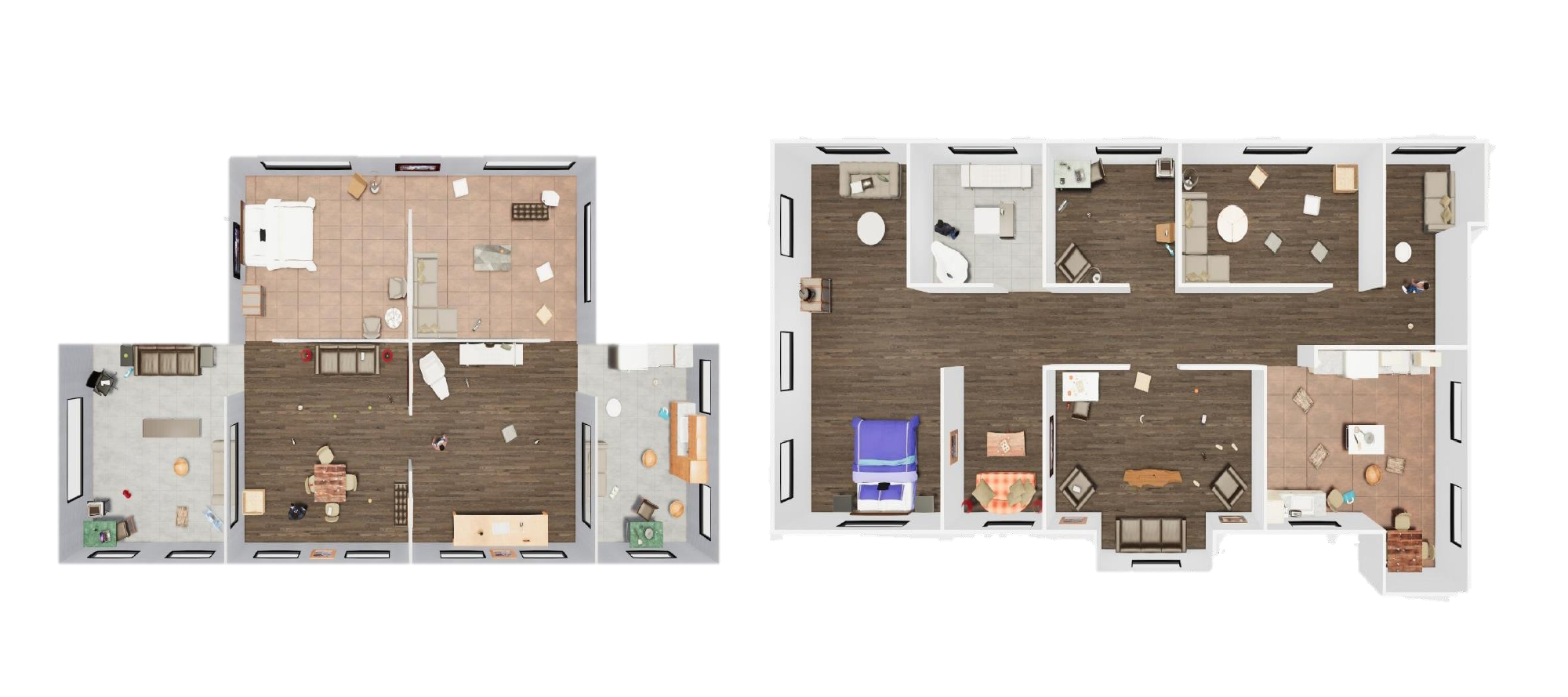}
    \caption{The example layouts of the TDW-MAT environments.}
\label{fig:tdw_env}
\end{figure}

\begin{table}[t!]
    \centering
    \begin{tabularx}{\columnwidth}{>{\raggedright\arraybackslash}p{0.25\columnwidth} >{\raggedright\arraybackslash}p{0.65\columnwidth}}
    \toprule
        Action & Description \\
     \midrule
        Move forward & move forward 0.5m \\
    \addlinespace
        Turn left & turn left by 15 degrees \\
    \addlinespace
        Turn right & turn right by 15 degrees \\
    \addlinespace
        Grasp & grasp an object \\
    \addlinespace
        Put In & put the target into the container \\
    \addlinespace
        Drop & drop the objects held in hand \\
    \addlinespace
        Send message & send a message to other agents \\
     \bottomrule
    \end{tabularx}
    \caption{The action space of the TDW-MAT environment.}
    \label{tab:tdw_action}
\end{table}

\subsection{TDW-MAT Environment}
TDW-MAT, an extended version of the ThreeDWorld Transport Challenge~\cite{Gan2021TheTT}, is built on  TDW~\cite{Gan2020ThreeDWorldAP}. It features more natural object placements and a variety of objects and containers that assist in transporting items. 
The \textit{common goal} involves transporting items in two categories: Food and Stuff. Each episode includes 10 target objects, and 2 to 5 containers, which are placed to facilitate moving multiple items simultaneously, as listed in Table~\ref{tab:tdw_obj}. 
Unlike C-WAH, agents in TDW-MAT cannot obtain complete room information without performing a 360-degree rotation in 15-degree increments. 
Communication is limited to 500 characters per frame, and $H$ is set to 3000 simulation steps. The layout is shown in Figure~\ref{fig:tdw_env} and the detailed action space is listed in Table~\ref{tab:tdw_action}.

\begin{figure}[t!]
    \centering 
    \includegraphics[width=\linewidth]{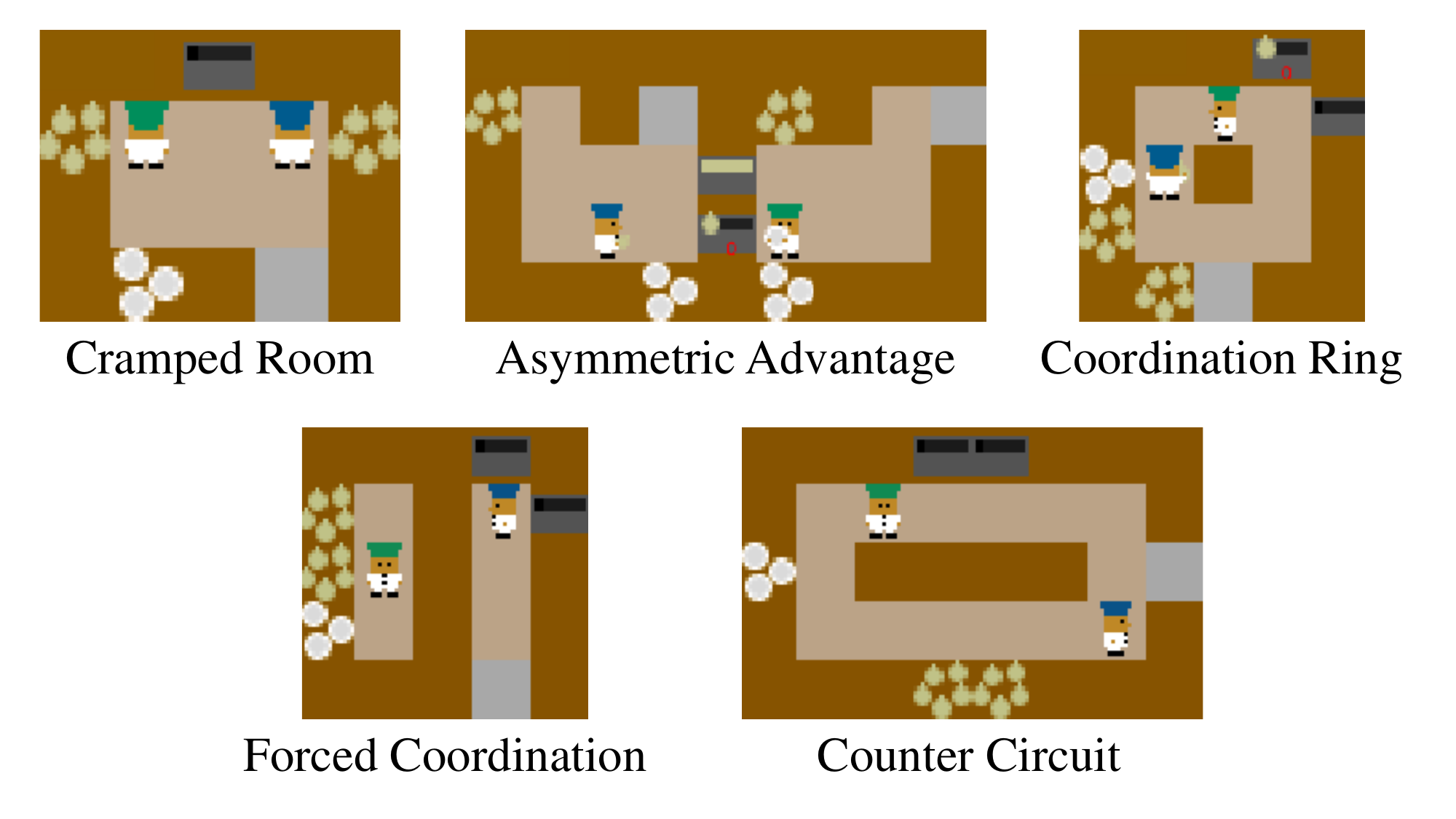}
    \caption{The example layouts of the Overcooked-AI environments.}
\label{fig:overcook_env}
\end{figure}

\begin{table}[t!]
    \centering
    \begin{tabularx}{\columnwidth}{>{\raggedright\arraybackslash}p{0.25\columnwidth} >{\raggedright\arraybackslash}p{0.65\columnwidth}}
    \toprule
        Action & Description \\
     \midrule
        North (Up) & move up to the adjacent tile \\
    \addlinespace
        South (Down) & move down to the adjacent tile \\
    \addlinespace
        East (Left) & move left to the adjacent tile \\
    \addlinespace
        West (Right) & move right to the adjacent tile \\
    \addlinespace
        Stay & remain in the current position without moving \\
    \addlinespace
        Pick Up & pick up an item (e.g., \textit{onion, dish}) from the environment \\
    \addlinespace
        Put In & put the object in hand into a pot \\
    \addlinespace
        Place Object & place the object in hand on a nearby counter or surface \\
    \addlinespace
        Fill Dish & fill a dish with soup once the soup is fully cooked \\
    \addlinespace
        Deliver Soup & deliver the soup to the serving area \\
     \bottomrule
    \end{tabularx}
    \caption{The action space of the Overcooked-AI environment.}
    \label{tab:overcooked_action}
\end{table}

\subsection{Overcooked-AI Environment}
Overcooked-AI~\cite{carroll2019utility} is a multiplayer game where two agents must cooperate to score as many points as possible. The cooking ingredients are swiftly prepared, placed into a pot, transferred to a dish, and served at the designated location, earning the agents 20 points for each completed dish. $H$ is set to 400 simulation steps, and unlike the three types of indoor multi-room simulation environments, Overcooked-AI is a fully observable environment, allowing the agents to access all information within the environment without the need for utterance actions. Overcooked-AI comprises five different layouts: Cramped Room, Asymmetric Advantage, Coordination Ring, Forced Coordination, and Counter Circuit. The layout is shown in Figure~\ref{fig:overcook_env} and the detailed action space is listed in Table~\ref{tab:overcooked_action}.

\subsection{Software and Hardware Setting}
We conducted the experiment on Ubuntu 22.04 LTS. The CPU used was an AMD Ryzen 7 5800X 8-Core Processor, the GPU was an NVIDIA GeForce RTX 3090 Ti with 24GB of VRAM, and the total memory was 32GB.

\section{Baseline Models and Versions of LLMs}
We conducted experiments that included all the state-of-the-art methods and the baseline models they employed. For C-WAH, TDW-MAT, and Noisy-C-WAH, the methods presented in the CoELA manuscript~\cite{zhang2023building} were used as baseline models, while for Overcooked-AI, the methods from the ProAgent manuscript~\cite{Zhang2023ProAgentBP} were used as baselines. Below, we provide detailed descriptions of each baseline model.

\subsection{For Partial Observable Experiment}
MHP, from the Watch-And-Help Challenge, is a Hierarchical Planner with a high-level planner utilizing MCTS and a low-level planner based on regression planning~\cite{Korf1987PlanningAS}. 

RHP, from the ThreeDWorld Transport Challenge, is a Hierarchical Planner with a high-level planner using heuristics and a low-level A-start-based planner for navigation, using a semantic map and Frontier Exploration strategy. 

CoELA leverages LLMs for planning and communication, enabling collaborative task-solving.
For a fair comparison and cost efficiency, we used \textit{gpt-4o-mini-2024-07-18} and \textit{gpt-3.5-turbo-0125} for the experiment and provided GPT-4-driven CoELA performance from the CoELA manuscript~\cite{zhang2023building} for a comprehensive analysis of different LLM capacities.

\subsection{For Fully Observable Experiment}
SP~\cite{tesauro1994td, carroll2019utility} involves agents learning by playing against copies of themselves, based on the TD($\lambda$) reinforcement learning algorithm.
    
PBT~\cite{jaderberg2017population} optimizes both the parameters and hyperparameters of a population of neural networks simultaneously. This approach enables the model to continuously improve by exploring different hyperparameter configurations rather than relying on a fixed set.

FCP~\cite{strouse2021collaborating} focuses on improving coordination between agents and unseen partners. FCP utilizes a training process where agents are tested against a diverse set of partners, ensuring improved generalization and better coordination when paired with novel agents or humans.

MEP~\cite{zhao2023maximum} introduces a Population Entropy bonus to encourage diversity among agents, aiming to reduce the negative effects of distributional shifts when collaborating with previously unseen partners, such as humans.

COLE~\cite{li2023cooperative, li2024tackling} addresses the challenge of zero-shot coordination by constructing open-ended objectives. COLE uses game theory and graph theory to evaluate and adapt to the strategies of diverse partners.

ProAgent~\cite{Zhang2023ProAgentBP} leverages LLMs for cooperative planning and inferring the intentions of collaborators. For a fair comparison and cost efficiency, we used \textit{gpt-4o-mini-2024-07-18} for the experiment.

BC~\cite{pomerleau1991efficient} is employed for the imitation of actions performed by human collaborators, and we used five different BC models in our experiments.

\subsection{Versions of LLMs}
REVECA leverages \textit{gpt-4o-mini-2024-07-18} through the OpenAI API, with the following parameters: a temperature of 0.7, top-p of 1, and a maximum token limit of 1024. This setup is also used by \textit{gpt-3.5-turbo-0125} and the open-source LLM, \textit{Meta-Llama-3.1-8B-Instruct}.

\begin{figure}[t!]
    \centering 
    \includegraphics[width=\linewidth]{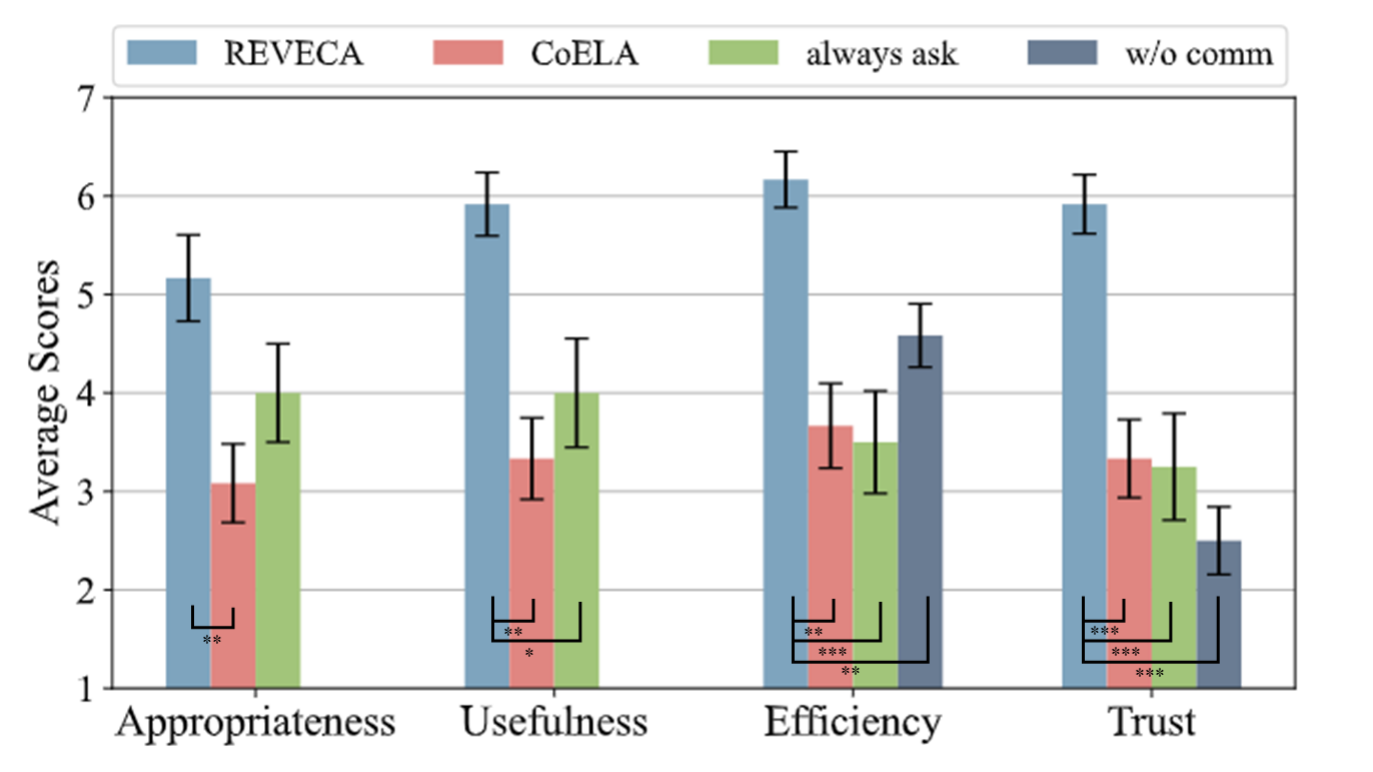}
    \caption{User study results with C-WAH. This figure illustrates the mean scores and associated standard errors for responses to four research questions. Statistical significance was denoted as * for \textit{p} $<$ 0.05, ** for \textit{p} $<$ 0.01, *** for \textit{p} $<$ 0.001.}
    \label{fig:supple_user_study_results}
\end{figure}

\section{Additional Results}
\subsection{User Study Results}
A paired t-test was conducted to compare the scores between REVECA and the other conditions in the user study, assessing the significance of any observed differences. 

The results are shown in Figure~\ref{fig:supple_user_study_results}. In Appropriateness, which evaluates whether the agent's responses align with the user's intent, significant differences were observed, except in the comparison between REVECA and the ``always ask'' condition. Furthermore, significant differences in Usefulness, Efficiency, and Trust were observed between REVECA and all other conditions.

\subsection{Inference Times}
\begin{table}[t!]
    \centering
    \begin{tabularx}{\columnwidth}{lcc}
    \toprule
        Component & REVECA & w/o CoT \\
     \midrule
        Relevance Estimation & 5.32 s & 0.75 s \\
        Adaptive Planning & 8.10 s & 0.94 s \\
        Trajectory-based Validation & 6.16 s & 0.73 s \\
        Communication & 1.25 s & 1.35 s \\
     \bottomrule
    \end{tabularx}
    \caption{Inference times for REVECA's components using GPT-4o-mini as the backbone.}
    \label{tab:Inference-time}
\end{table}

The superior reasoning performance of LLMs is often attributed to their high computational cost. However, as the computational demands increase, employing LLMs in embodied agents can introduce communication and processing delays that may affect real-time performance. To evaluate this, we measured the inference times of REVECA's components in seconds using GPT-4o-mini as the backbone. Additionally, to analyze the extra inference time introduced by the Chain-of-Thought (CoT) reasoning, we conducted experiments with a version of REVECA without CoT (w/o CoT).

The experimental results are summarized in Table \ref{tab:Inference-time}. The original REVECA with CoT required 5.32 seconds for Relevance Estimation, 8.10 seconds for Adaptive Planning, and 6.16 seconds for Trajectory-based Validation. Communication, which did not involve CoT, demanded a similar time for both the CoT-enabled and w/o CoT versions. By disabling CoT, REVECA reduced the inference time for each component to approximately 1 second. This configuration meets real-time operational requirements, enabling agents to respond promptly during communication or task execution. Although disabling CoT slightly increased the number of Simulation Steps (from 44.2 to 48.9, as discussed in our paper), it still outperformed previous methods utilizing GPT-4 (Simulation Steps: 57.0). The trade-off between performance and speed introduced by CoT suggests that disabling CoT may be a viable option depending on the intended application.

Furthermore, the field of LLMs is advancing rapidly, with emerging models offering faster inference and reduced computational requirements. REVECA's compatibility with various LLM backbones ensures that its real-time performance will continue to improve as more efficient models become available.

\section{Key processes of REVECA}

\subsection{Message Generation and Extraction}
REVECA's Communication Module, which facilitates natural language information sharing, is invoked in four cases. 1) Simulation Initiation: Agents exchange initial information about their locations and surrounding objects. 2) Validation Requests: Agents query about task history. 3) Response to Validation Requests: Agents provide task completion history. 4) Sub-goal Achievement: Agents announce sub-goal completion.

\begin{figure*}[h!]
    \centering 
    \includegraphics[width=0.8\linewidth]{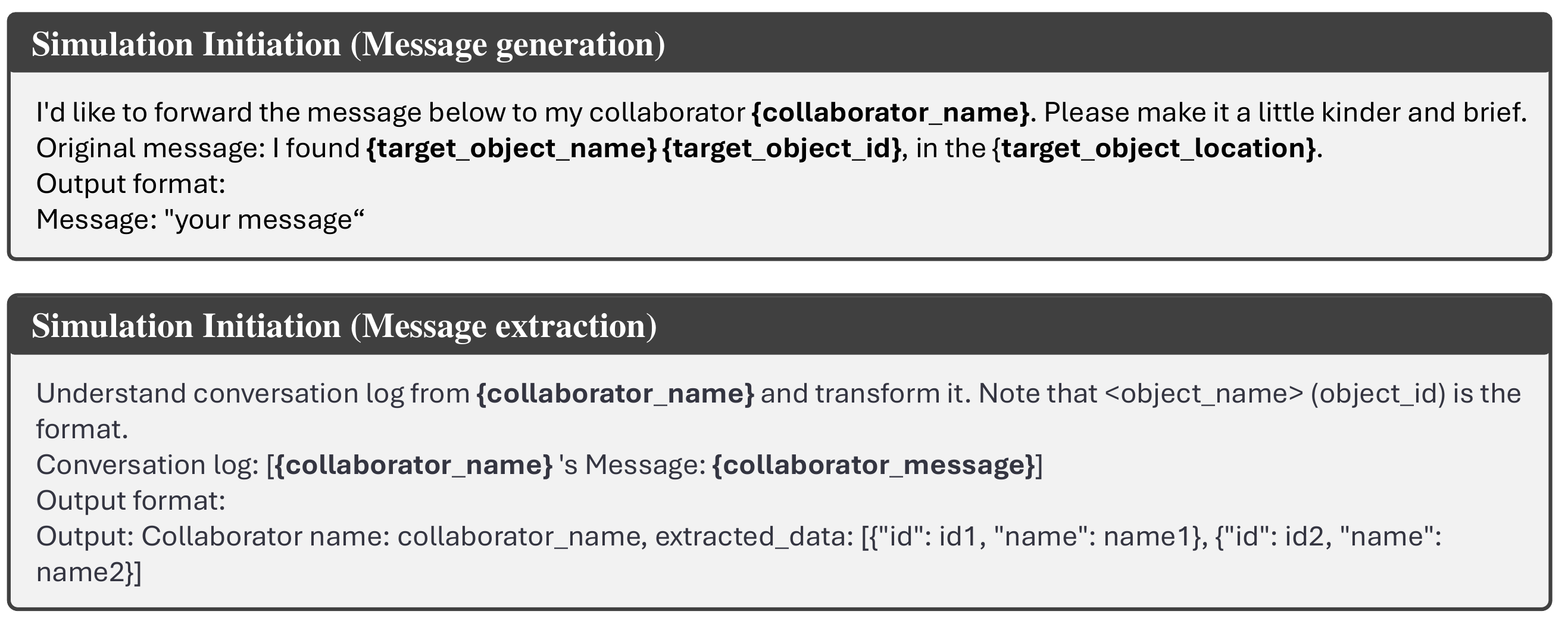}
    \caption{Prompt template for Simulation Initiation.}
\label{fig:init_prompt_template}
\end{figure*}

\begin{figure*}[h!]
    \centering 
    \includegraphics[width=0.8\linewidth]{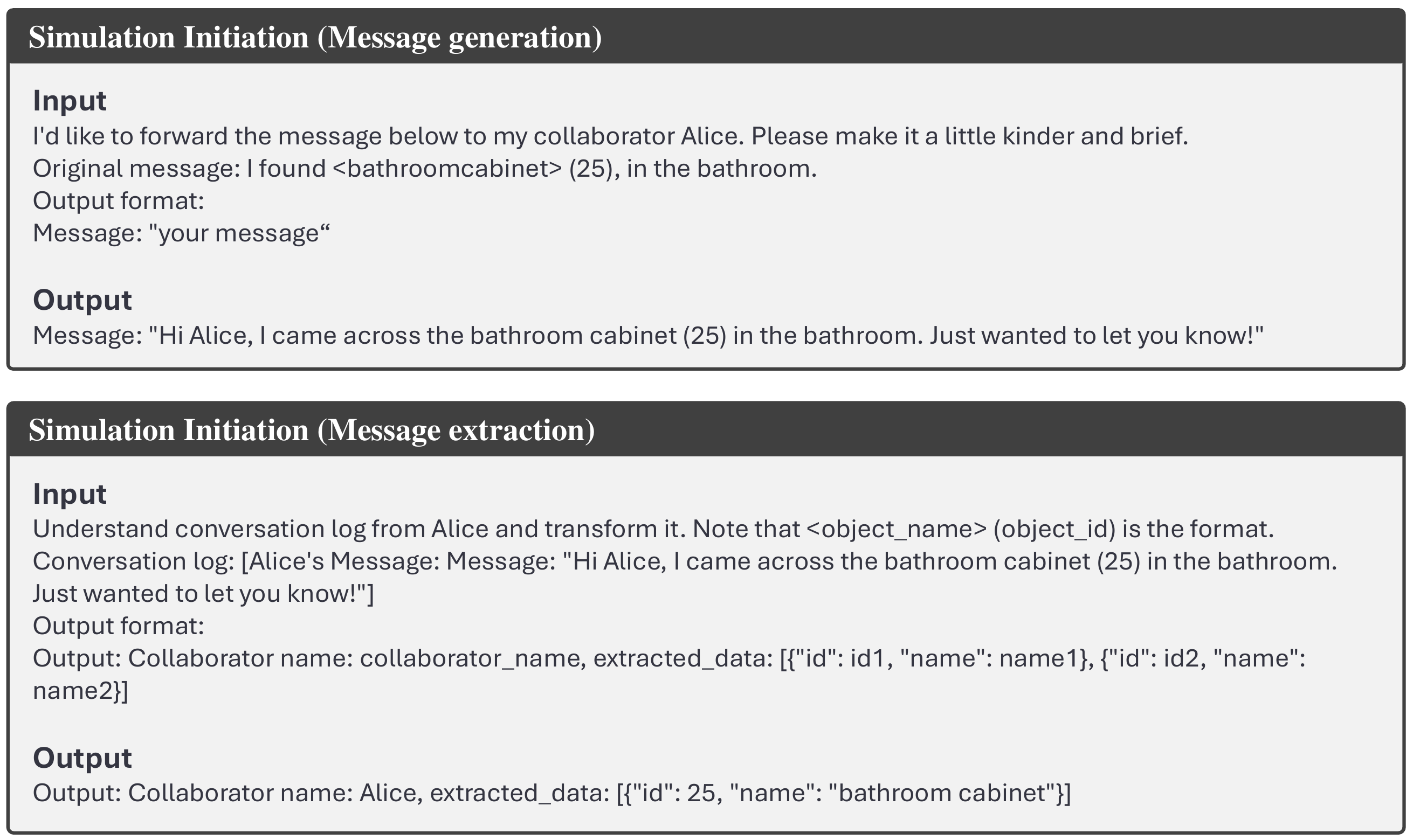}
    \caption{Actual example for Simulation Initiation.}
\label{fig:init_prompt_example}
\end{figure*}

\begin{figure*}[h!]
    \centering 
    \includegraphics[width=0.8\linewidth]{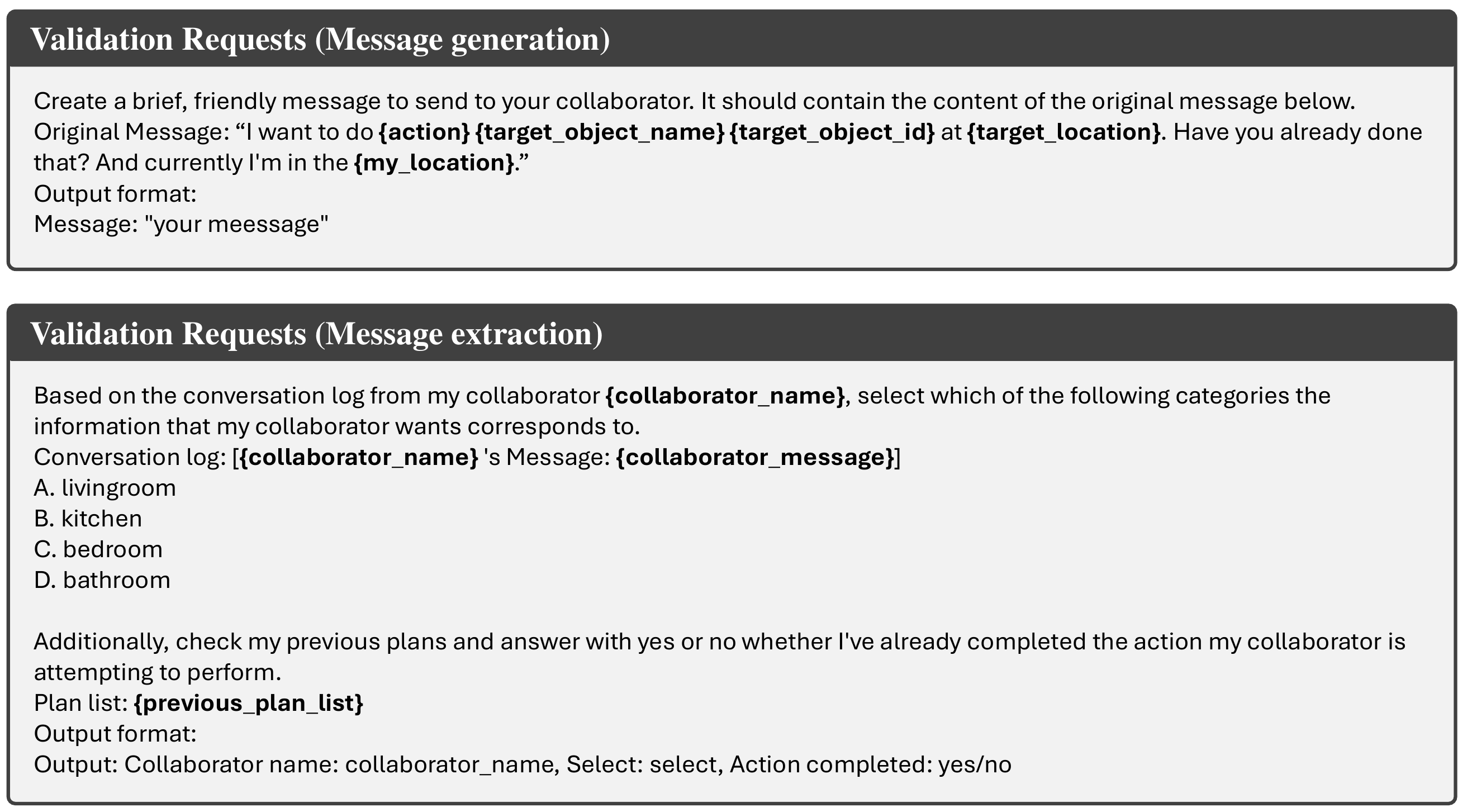}
    \caption{Prompt template for Validation Requests.}
\label{fig:request_prompt_template}
\end{figure*}

\begin{figure*}[h!]
    \centering 
    \includegraphics[width=0.8\linewidth]{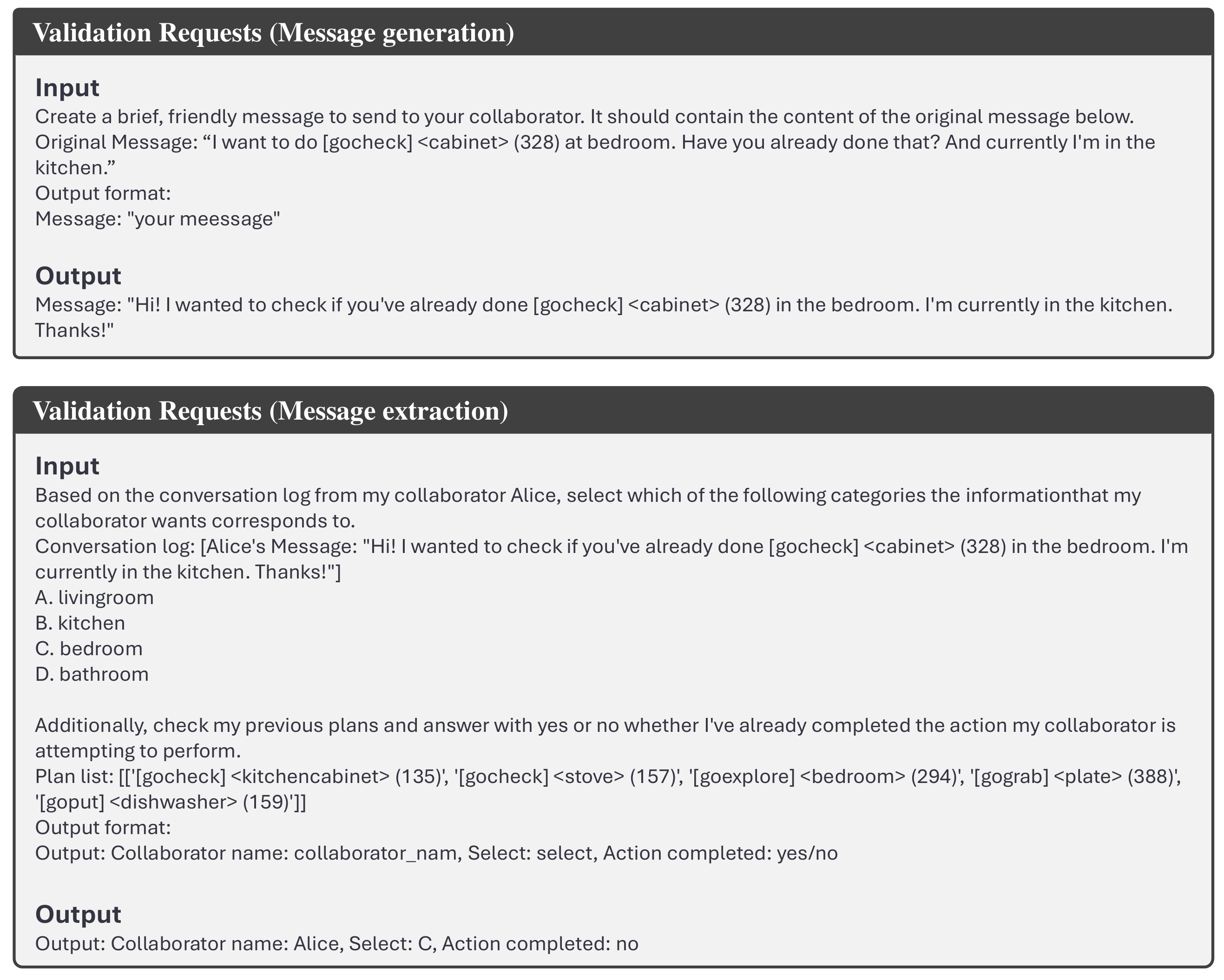}
    \caption{Actual example for Validation Requests.}
\label{fig:Validation_Requests_prompt_example}
\end{figure*}

\begin{figure*}[h!]
    \centering 
    \includegraphics[width=0.8\linewidth]{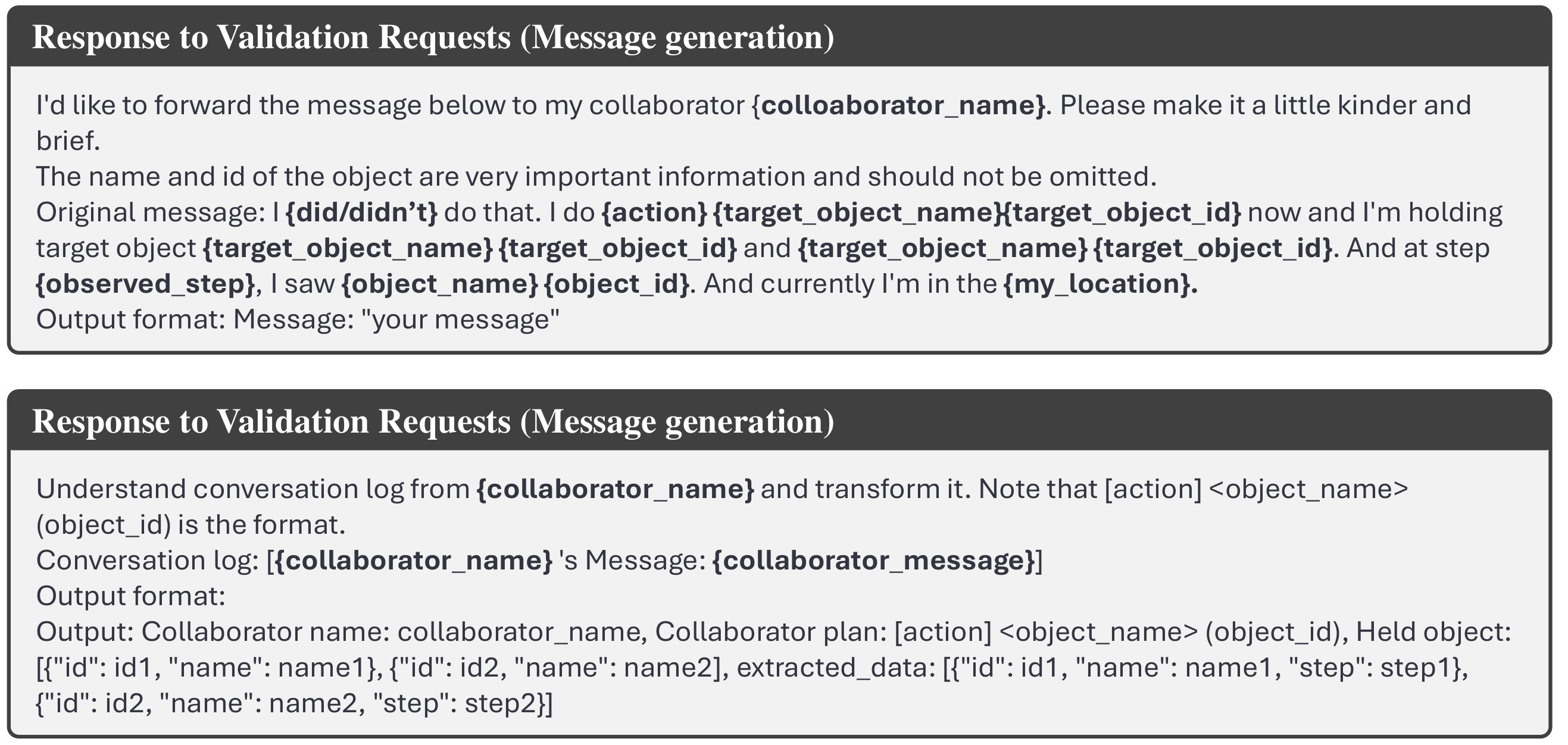}
    \caption{Prompt template for Response to Validation Requests.}
\label{fig:response_prompt_template}
\end{figure*}

\begin{figure*}[h!]
    \centering 
    \includegraphics[width=0.8\linewidth]{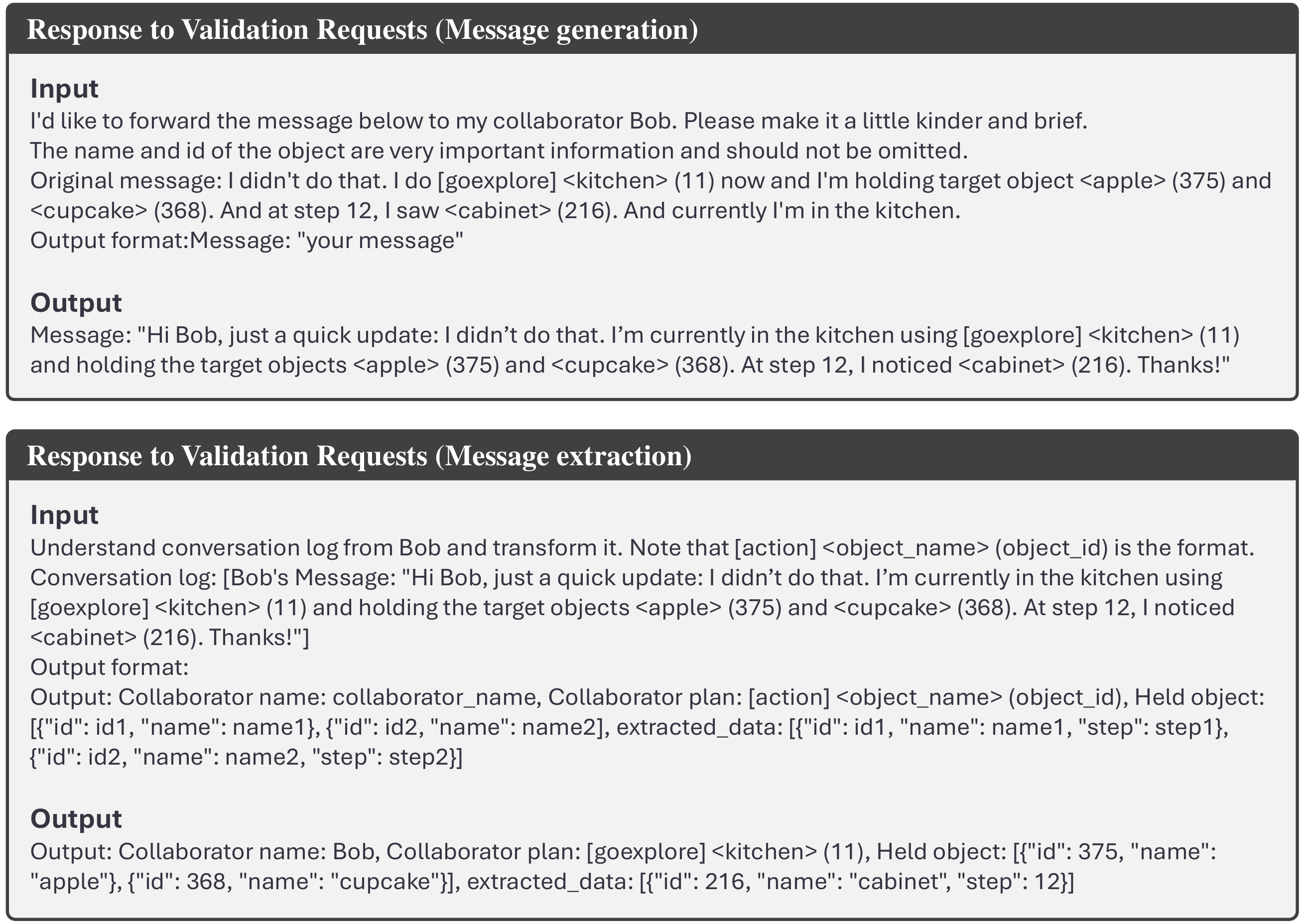 }
    \caption{Actual example for Response to Validation Requests.}
\label{fig:Response_prompt_example}
\end{figure*}

\begin{figure*}[h!]
    \centering 
    \includegraphics[width=0.8\linewidth]{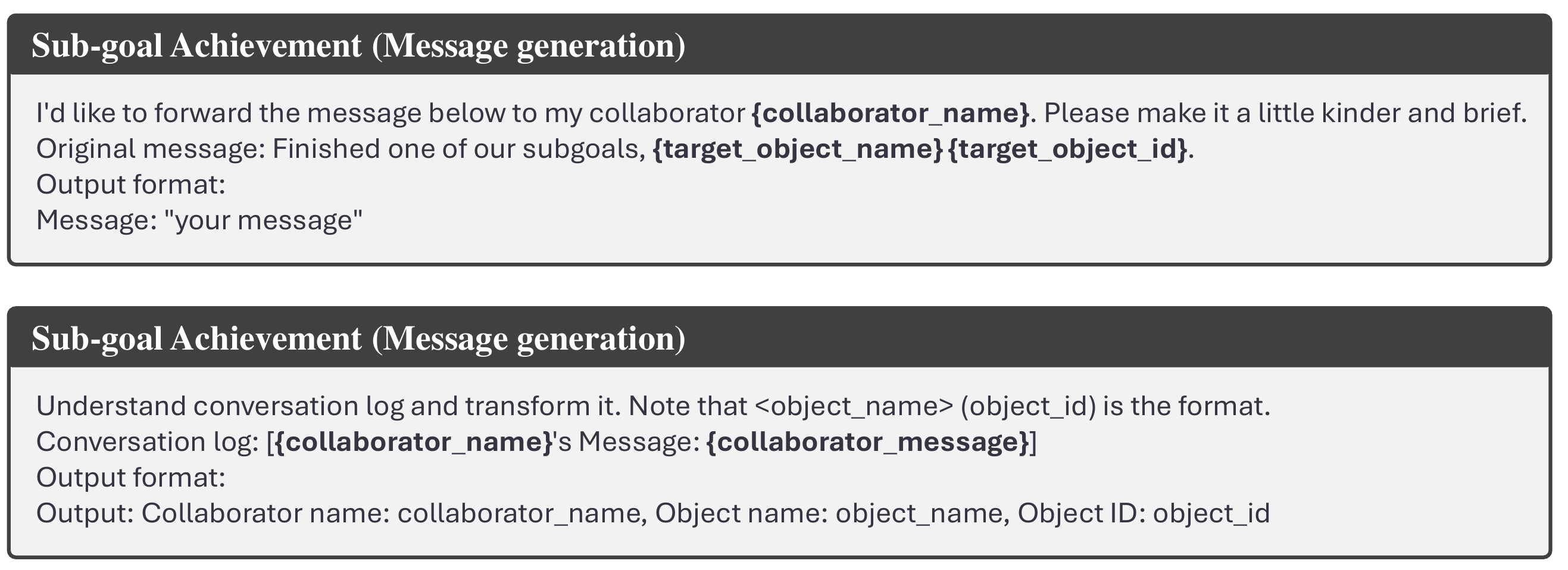}
    \caption{Prompt template for Sub-goal Achievement.}
\label{fig:subgoal_prompt_template}
\end{figure*}

\begin{figure*}[h!]
    \centering 
    \includegraphics[width=0.8\linewidth]{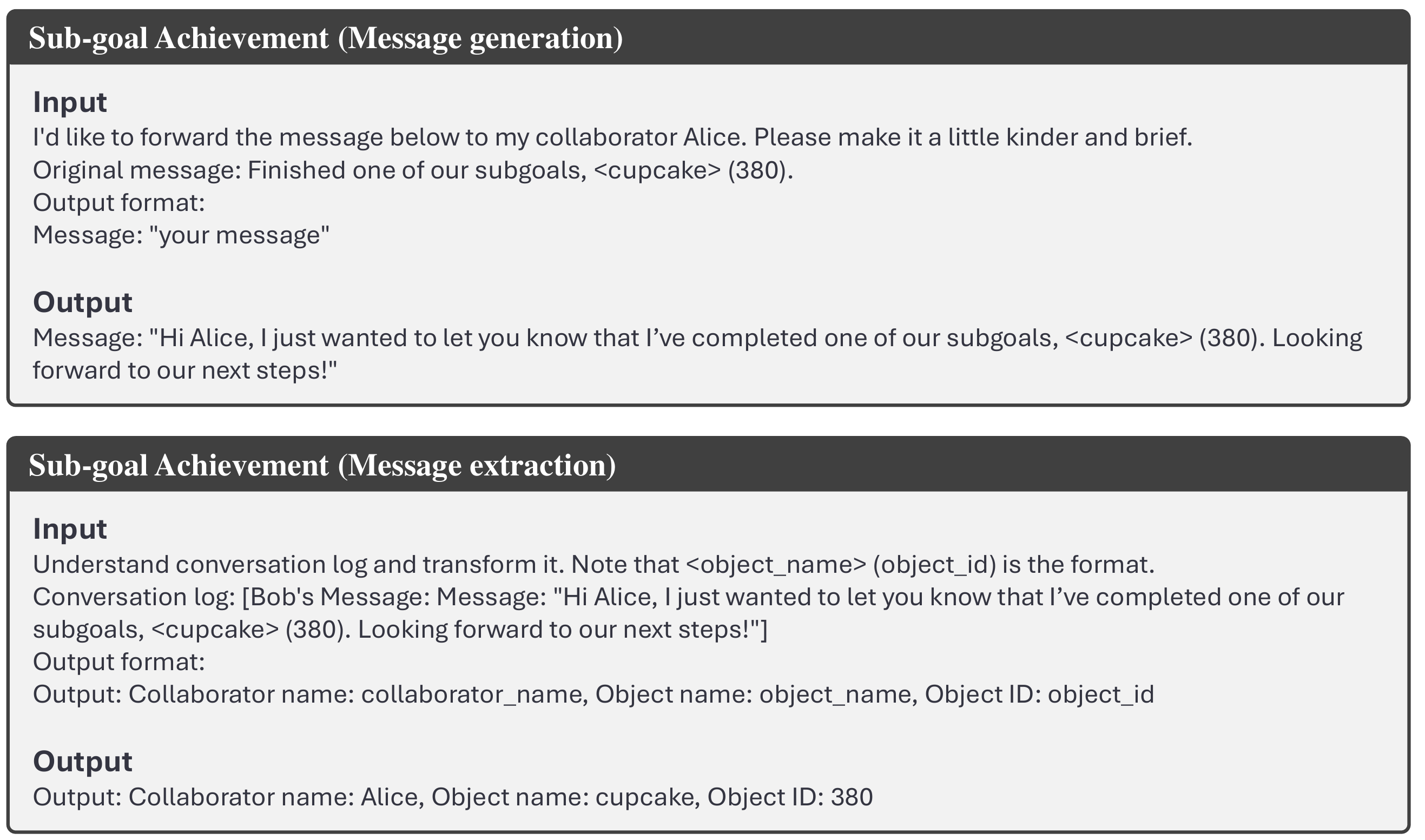}
    \caption{Actual example for Sub-goal Achievement.}
\label{fig:Sub-goal_prompt_example}
\end{figure*}

To generate contextually appropriate messages regardless of the scale of the LLMs, we employed different prompts for each case. After generating an initial rule-based message, we structured the prompt in a way that allows the LLMs to refine it into a more natural message. This approach enables REVECA to function effectively across various LLMs. 
The prompt template and an actual example of Simulation Initiation are illustrated in Figure~\ref{fig:init_prompt_template} and ~\ref{fig:init_prompt_example}, respectively. Similarly, the prompt template and actual example of Validation Requests are shown in Figure~\ref{fig:request_prompt_template} and ~\ref{fig:Validation_Requests_prompt_example}, Response to Validation Requests in Figure~\ref{fig:response_prompt_template} and ~\ref{fig:Response_prompt_example}, and Sub-goal Achievement in Figure~\ref{fig:subgoal_prompt_template} and~\ref{fig:Sub-goal_prompt_example}. These figures can be found in the Prompt Template and Actual Example section.



\subsection{Relevance Estimation}
\begin{figure*}[h!]
    \centering 
    \includegraphics[width=0.8\linewidth]{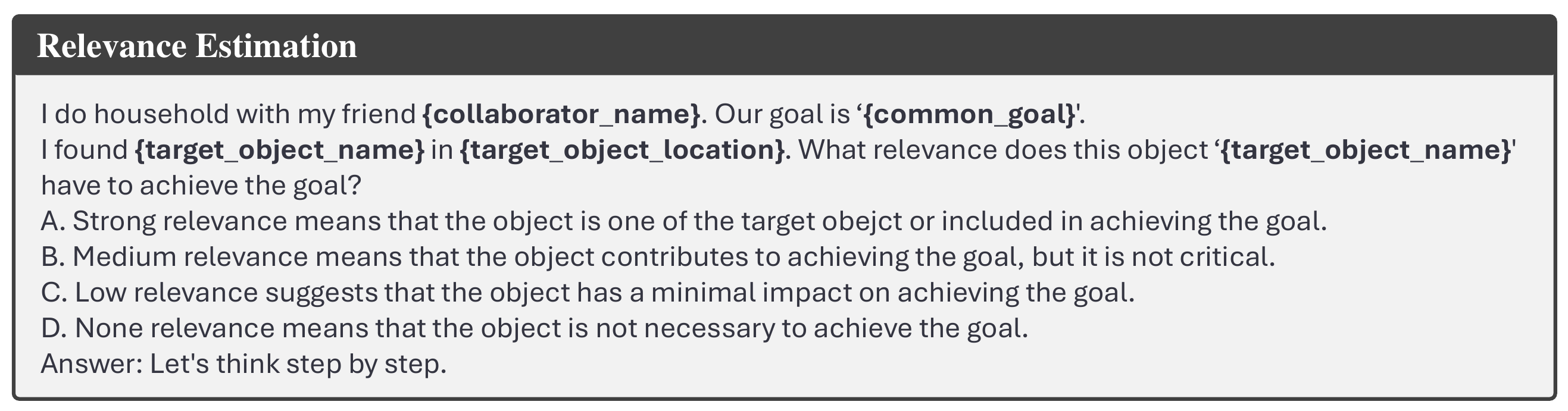}
    \caption{Prompt template for Relevance Estimation.}
\label{fig:relevance_prompt_template}
\end{figure*}

\begin{figure*}[h!]
    \centering 
    \includegraphics[width=0.8\linewidth]{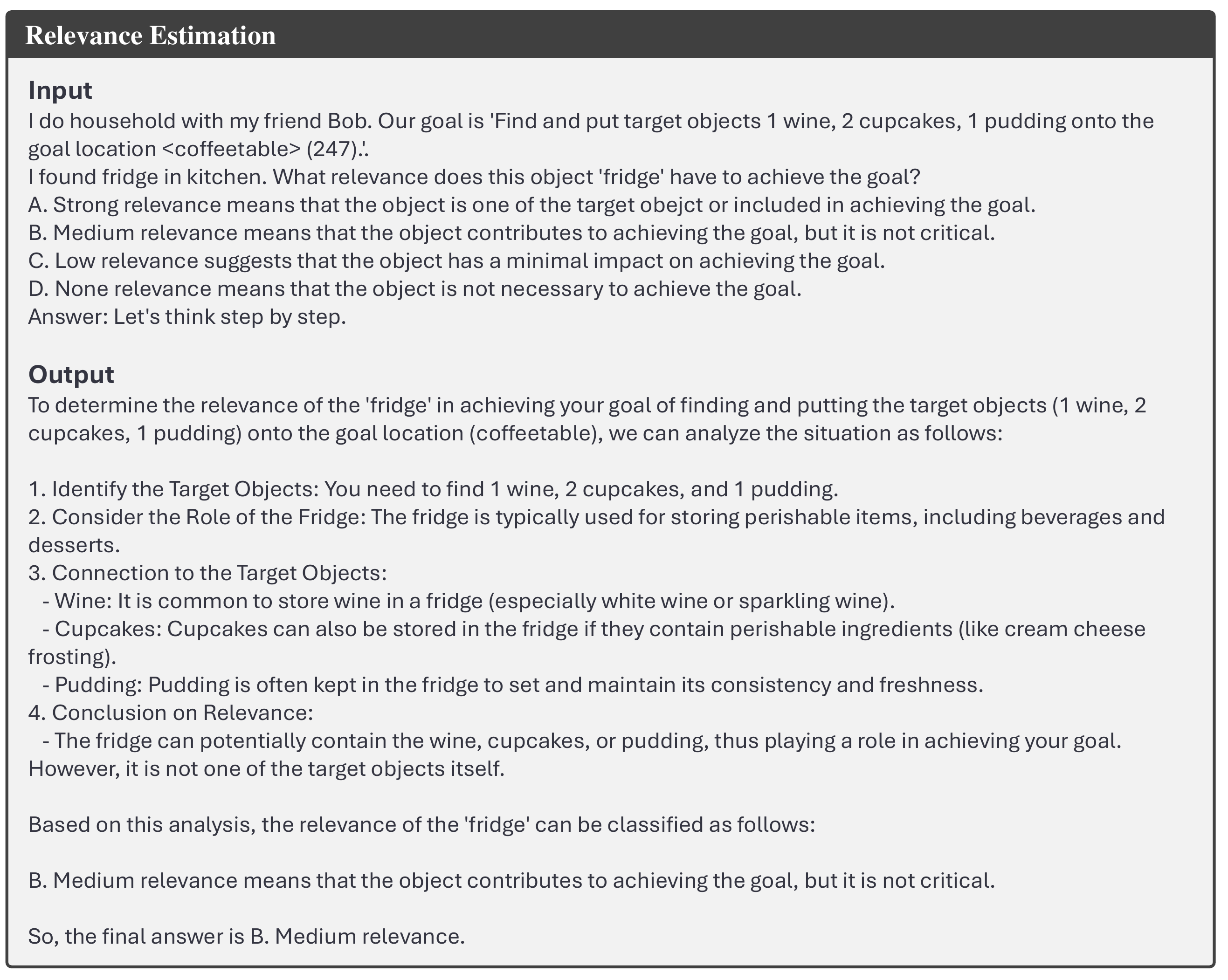}
    \caption{Actual example for Relevance Estimation.}
\label{fig:Relevance_prompt_example}
\end{figure*}

Based on the acquired $I_o$, REVECA evaluates the relevance score $R$ of each object in relation to achieving the $G$. The relevance score $R$ consists of four levels: \{Strong, Medium, Low, None\}, and is assessed by LLMs. A sample prompt template is illustrated in Figure~\ref{fig:relevance_prompt_template}, and an actual example is shown in Figure~\ref{fig:Relevance_prompt_example}. These figures can be found in the Prompt Template and Actual Example section.

\begin{algorithm}[t!]
\caption{Retrieving a set of $K$ information ($K=3$)}
\label{alg:algorithm}
\textbf{Input}: $M_o = \{I_o^1, I_o^2, \dots, I_o^n\}$ ($n$: number of information) \\
\textbf{Output}: Top $K$ information sorted by $R_i$ in descending order and by $P_i$ in descending order for ties.

\begin{algorithmic}[1]
\FOR{$i = 1$ to $n-1$}
    \FOR{$j = i + 1$ to $n$}
        \IF{$R_i < R_j$ \OR ($R_i = R_j$ \AND $P_i < P_j$)}
            \STATE Swap $I_o^i$ and $I_o^j$
        \ENDIF
    \ENDFOR
\ENDFOR
\STATE \textbf{return} $\{I_o^1, I_o^2, \dots, I_o^K\}$
\end{algorithmic}
\end{algorithm}

\subsection{Retrieving a Set of $K$ Information}
To retrieve $K$ pieces of $I_o$, REVECA utilizes the relevance score $R$ and the relative proximity $P$. $R$ is the score measured during Observation Time, while $P$ is calculated based on the positions of $I_o$ and $I_c$ during Planning Time. Once $P$ has been calculated, all instances of $I_o$ stored in $M_o$ are sorted in descending order based on $R$ and $P$. Through this process, we obtain the top $K$ pieces of information. This process is summarized in Algorithm~\ref{alg:algorithm}.

\subsection{Adaptive Planning}
\begin{figure*}[h!]
    \centering 
    \includegraphics[width=0.8\linewidth]{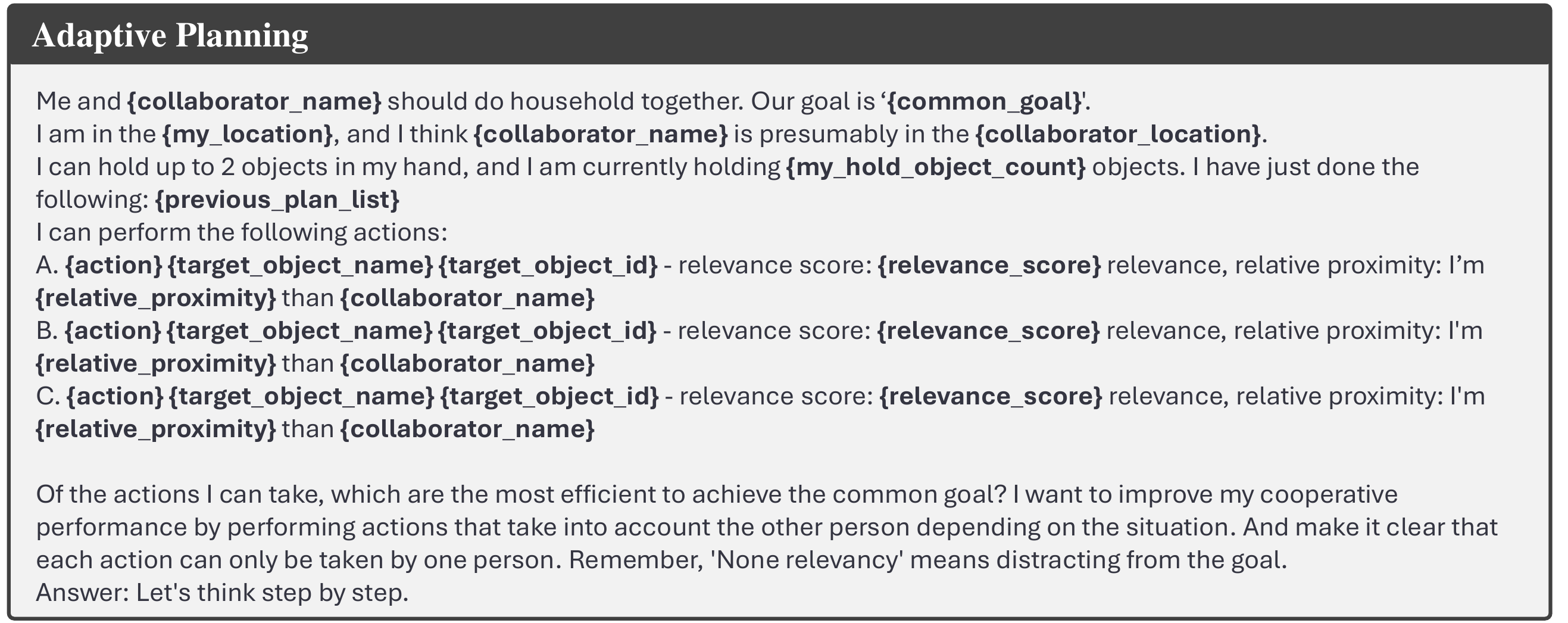}
    \caption{Prompt template for Adaptive Planning.}
\label{fig:planning_prompt_template}
\end{figure*}

\begin{figure*}[h!]
    \centering 
    \includegraphics[width=0.8\linewidth]{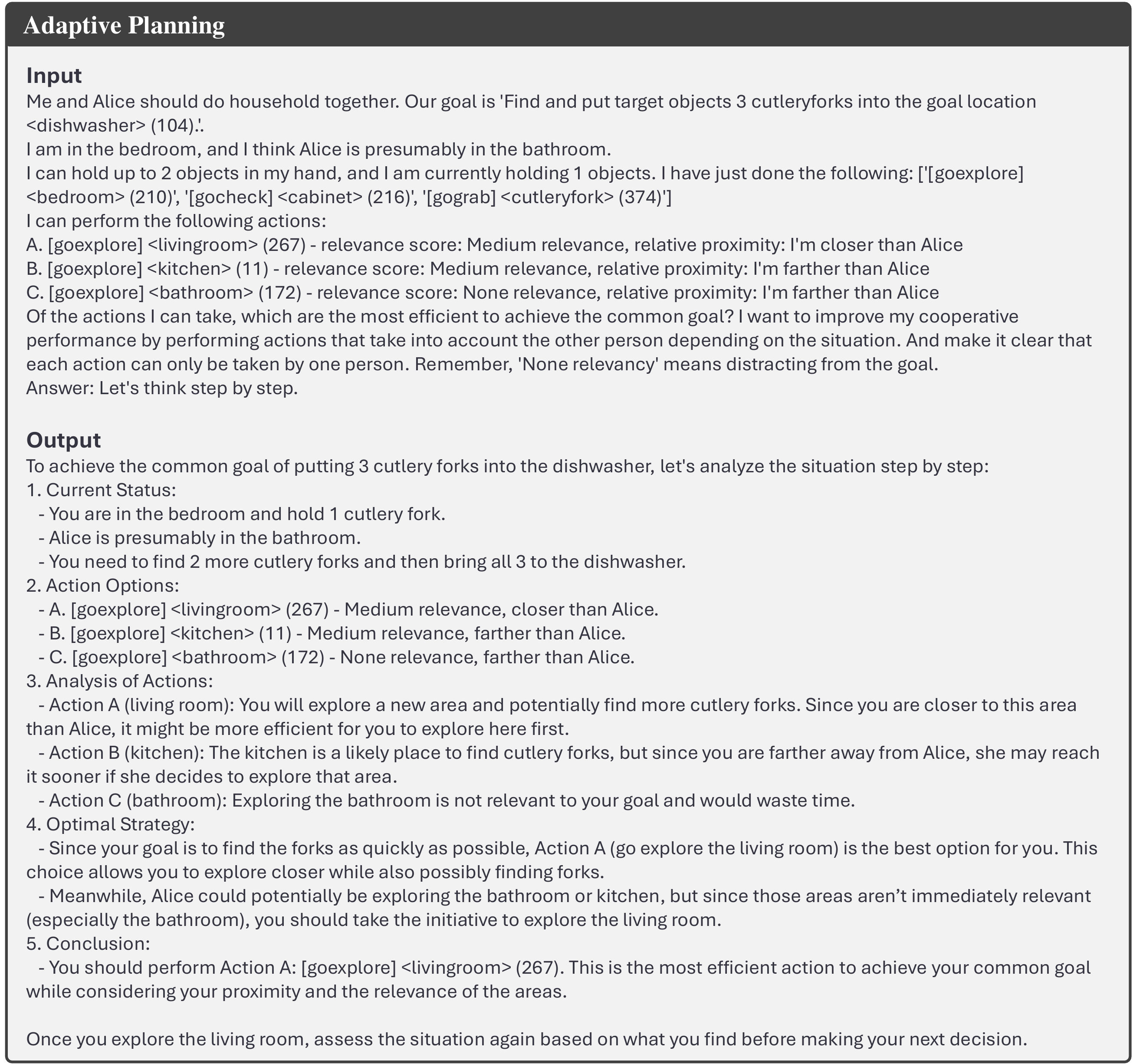}
    \caption{Actual example for Adaptive Planning.}
\label{fig:planning_prompt_example}
\end{figure*}

For LLMs-based planning, not only the top $K$ pieces of $I_o$, $R$, and $P$, but also the agent’s current information, such as held objects, current position, and completed plan history are incorporated into the prompt as additional context.
An example of the prompt template used for planning is illustrated in Figure~\ref{fig:planning_prompt_template} and an actual example is shown in Figure~\ref{fig:planning_prompt_example}. These figures can be found in the Prompt Template and Actual Example section.

\subsection{Trajectory-based Validation}
\begin{figure*}[h!]
    \centering 
    \includegraphics[width=0.8\linewidth]{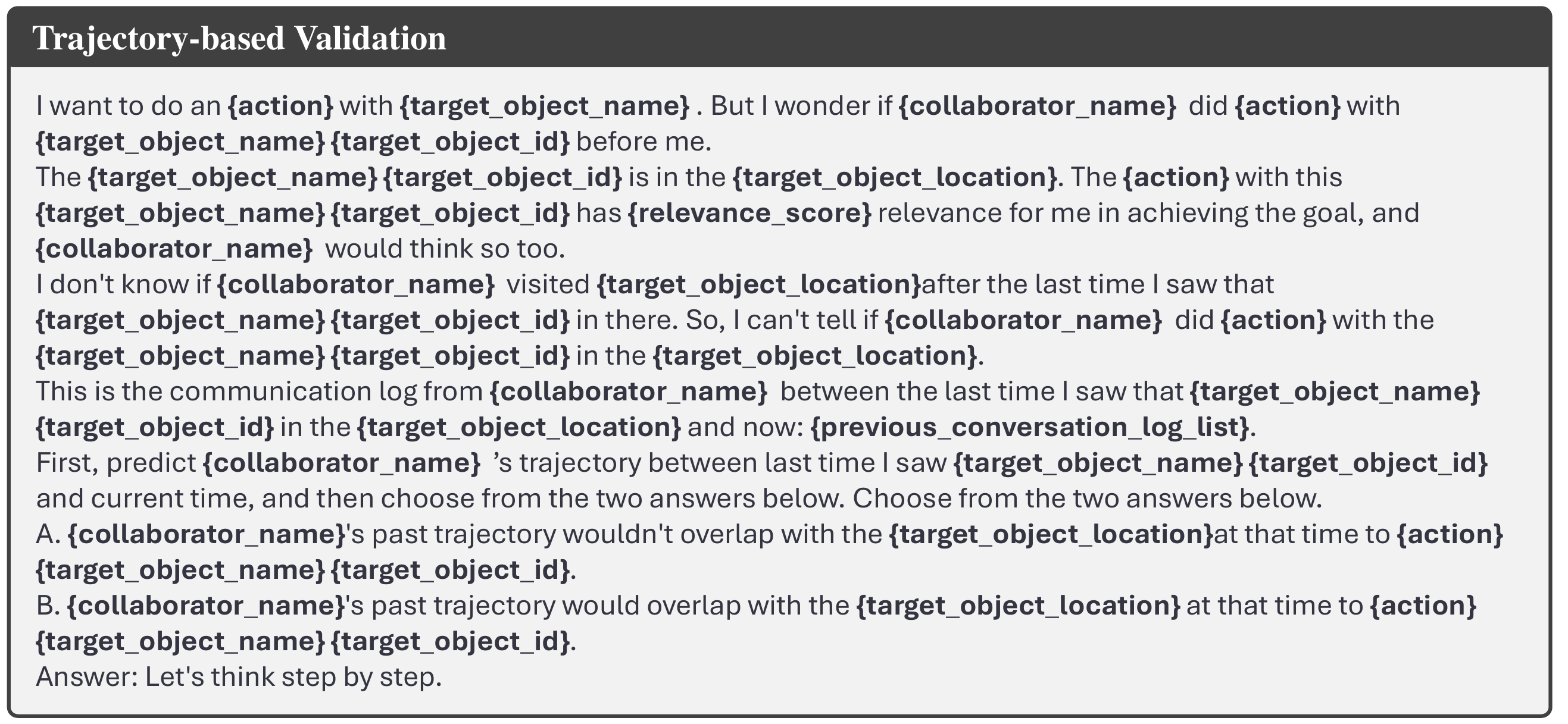}
    \caption{Prompt template for Trajectory-based Validation.}
\label{fig:validation_prompt_template}
\end{figure*}

\begin{figure*}[h!]
    \centering 
    \includegraphics[width=0.8\linewidth]{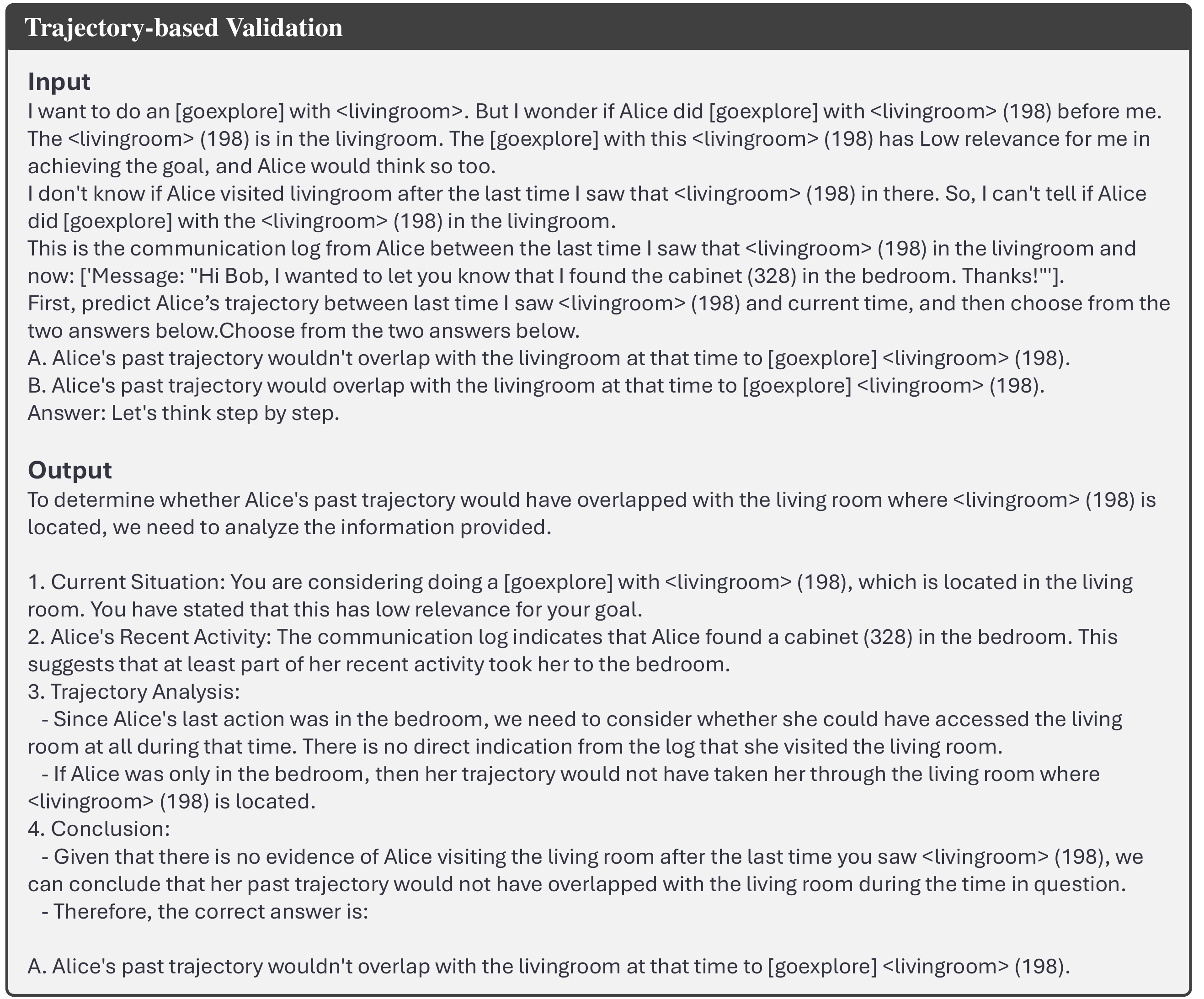}
    \caption{Actual example for Trajectory-based Validation.}
\label{fig:Trajectory_prompt_example}
\end{figure*}

For Trajectory-based Validation, the $I_o$ and $R$ used in planning, the generated plan, and $I_c$ are included in the prompt. 
An example of the prompt template is illustrated in Figure~\ref{fig:validation_prompt_template} and an actual example is shown in Figure~\ref{fig:Trajectory_prompt_example}. These figures can be found in the Prompt Template and Actual Example section.

\section{Further Discussions in Complex Environments}
\subsection{Beyond Two-Agent Scenarios}
While this paper primarily focuses on a two-agent collaboration involving Alice and Bob, the REVECA framework is inherently scalable and applicable to multi-agent settings with three or more agents. Communication Module and Relevance Estimation process are designed to be invariant with respect to the number of agents. The mechanisms for top-k information retrieval and Adaptive Planning remain largely consistent across varying agent counts; however, increasing the number of agents leads to a higher $I_c$ and an overall increase in the volume of information in the Memory Module. Consequently, the application of top-k sampling based on relevance scores might need a more granular scale than the current four-level scheme. Furthermore, the computational intensity and processing time of the Trajectory-based Validation module may scale linearly with the number of agents, potentially necessitating more efficient strategies.

\subsection{Open-Ended Environments}
In our experimental setup, Common goal $G$ is explicitly defined at the start of the simulation and remain static throughout its duration. Scenarios involving changing objectives or critical subtask order dependencies necessitate further advancements in both relevance estimation and adaptive planning. The current validation process might encounter limitations if agents are unable to accurately identify sources of outdated information beyond their observational range. We defer these challenges to future research.

\onecolumn 

\section{Prompt Template and Actual Example}

\subsection{Conversation Prompts: Simulation Initiation}

\newpage
\clearpage
\subsection{Conversation Prompts: Validation Requests}

\newpage
\clearpage
\subsection{Conversation Prompts: Response to Validation Requests}

\newpage
\clearpage
\subsection{Conversation Prompts: Sub-goal Achievement}

\newpage
\clearpage
\subsection{Relevance Estimation Prompts}

\newpage
\clearpage
\subsection{Adaptive Planning Prompts}

\newpage
\clearpage
\subsection{Trajectory-based Validation Prompts}

\onecolumn 
\newpage
\twocolumn 

\section{Example Scenarios}
\subsection{Efficient Memory Management}
In noisy environments, agents often accumulate irrelevant information, which overwhelms their memory and impairs decision-making. Our Relevance Estimation mechanism enables agents to prioritize goal-relevant data, effectively avoiding distractions from irrelevant objects.

Consider the agent Alice, who is assigned the goal: "Find and put target objects 1 wine, 2 cupcakes, 1 pudding onto the goal location $<$coffeetable$>$ (247)." If we encounter a fork during the process, it is intuitively evident that the fork is unrelated to the goal, as the relevant objects for the goal are wine, cupcakes, pudding, and the goal location coffeetable.

Similarly, REVECA employs Relevance Estimation to evaluate the relevance of acquired information before proceeding to the Planning Time. This approach addresses the significant overhead in planning caused by previous methods, which either utilized all accumulated information or only the most recent $K$ pieces of information.

\begin{figure*}[h!]
    \centering 
    \includegraphics[width=0.8\linewidth]{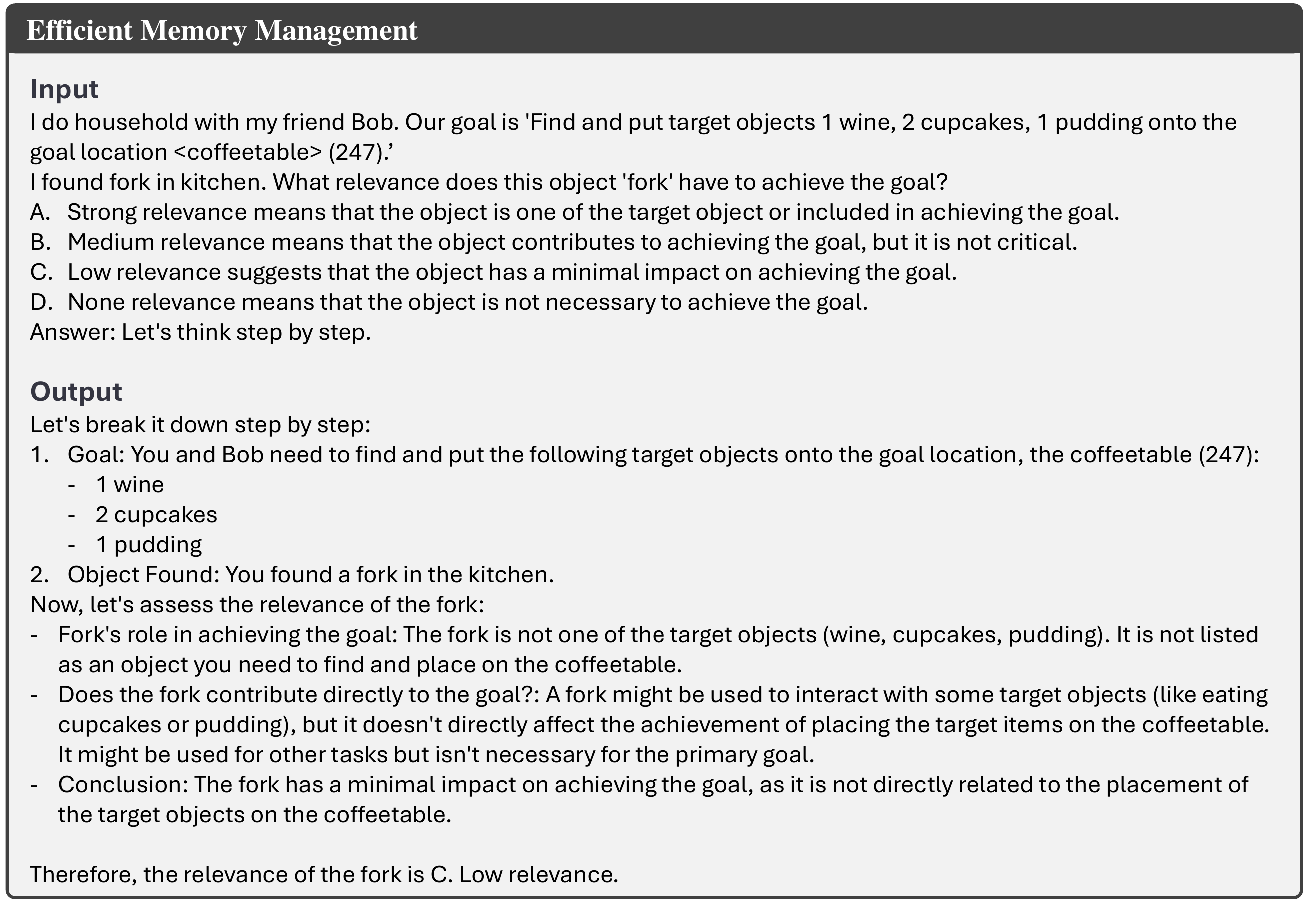}
    \caption{Prompt example for efficient memory management.}
\label{fig:efficient_memory_management}
\end{figure*}

In Figure~\ref{fig:efficient_memory_management}, the agent Alice acquires information about the fork from the environment and evaluates its relevance to the goal. Alice infers that the fork is not among the target objects and does not directly contribute to goal achievement. Consequently, Alice assigns a $Low Relevance$ score to the fork. This process reduces the likelihood of the fork influencing the planning phase, as the planning algorithm prioritizes information with high relevance scores.

\subsection{Globally Optimal Planning}
In scenarios where multiple collaborators work together to achieve a common goal, failure to consider the positions of other collaborators may lead agents to make decisions resulting in sub-optimal collaboration. Unlike previous studies, which often overlooked this issue, REVECA enables the generation of optimal plans through Adaptive Planning.

For example, consider the early stage of a simulation where agent Alice must decide which room to explore next. Depending on the current positions of Alice and Bob, some rooms may be closer to Alice, while others may be closer to Bob. Since the shared objective of both agents is to achieve the given goal as quickly as possible, agent Alice needs to consider Bob's current location when generating her optimal plan. Intuitively, Alice would determine that it is optimal for her to explore the room closer to her.

\begin{figure*}[h!]
    \centering 
    \includegraphics[width=0.8\linewidth]{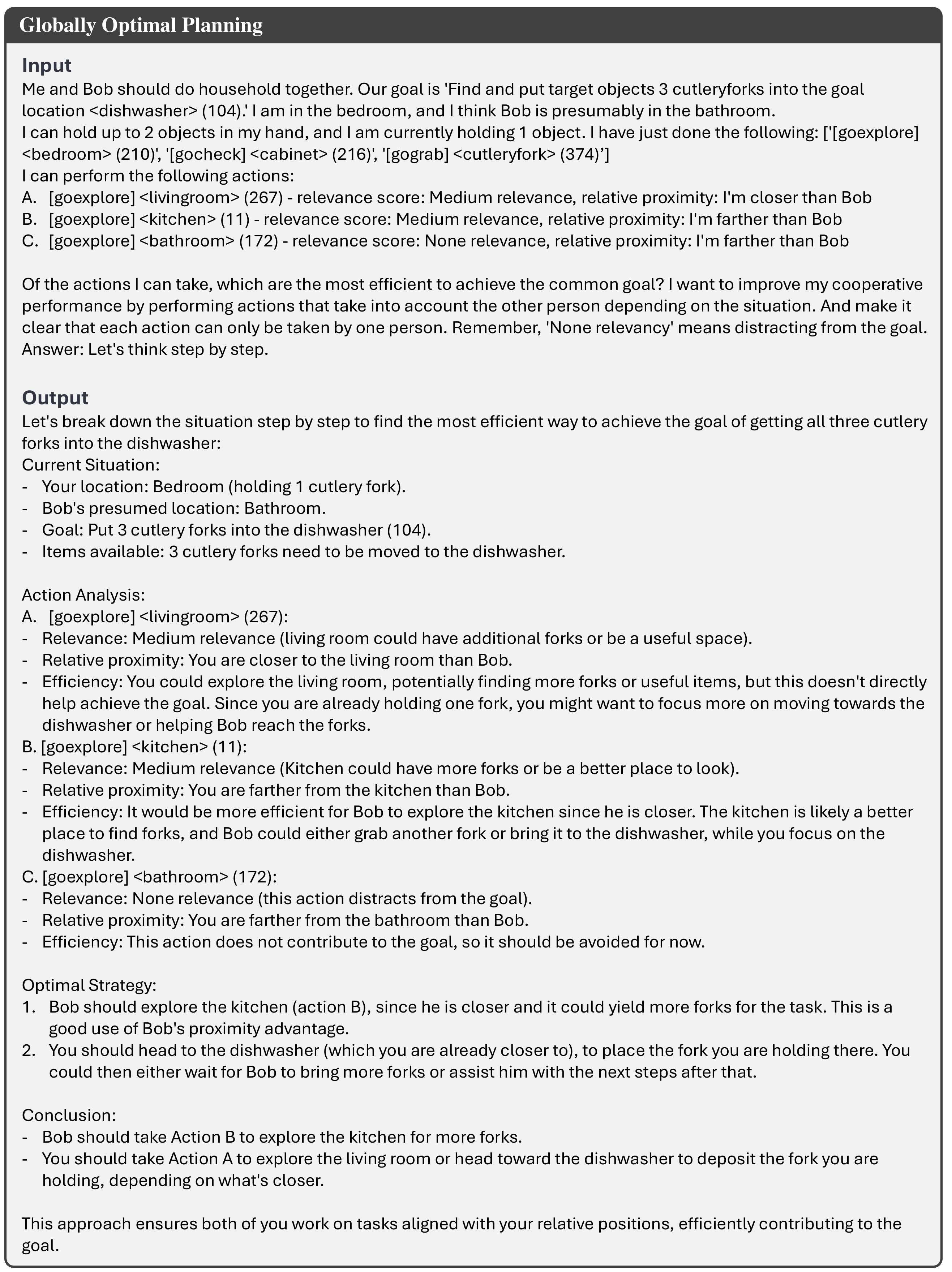}
    \caption{Prompt example for globally optimal planning.}
\label{fig:globally_optimal_planning}
\end{figure*}

In Figure~\ref{fig:globally_optimal_planning}, agent Alice faces a scenario where she can explore the living room, kitchen, or bathroom. Alice is provided with the Relevance Score of each plan, indicating its relevance to the goal, as well as the Relative Proximity of the rooms to either agent. Based on this information, Alice determines that the bathroom is irrelevant to the goal, and the kitchen is closer to Bob. Thus, Alice concludes that exploring the living room is the optimal plan and selects Action A as her next step. By choosing the next action in this manner, each agent increases the likelihood of generating an optimal plan based on the most efficient paths.

\subsection{Prevention of False Planning}
In partially observable environments, agents' knowledge can become outdated due to unseen interactions with collaborators, leading to false plans. A false plan may force an agent to make meaningless moves, which can significantly reduce collaboration performance.

Suppose that during the Planning Time, Alice’s next action is selected as [gograb] $<$cupcake$>$ (216). At this point, Alice is not in the kitchen where the cupcake is located, and time has passed since she last acquired this information. Since Bob evaluated the cupcake as highly relevant like Alice did, he may have already grabbed it if he is in the kitchen. If Bob has already taken the cupcake, Alice's [gograb] $<$cupcake$>$ (216) would be a false plan. It is crucial to prevent this action from being executed beforehand.

\begin{figure*}[h!]
    \centering 
    \includegraphics[width=0.8\linewidth]{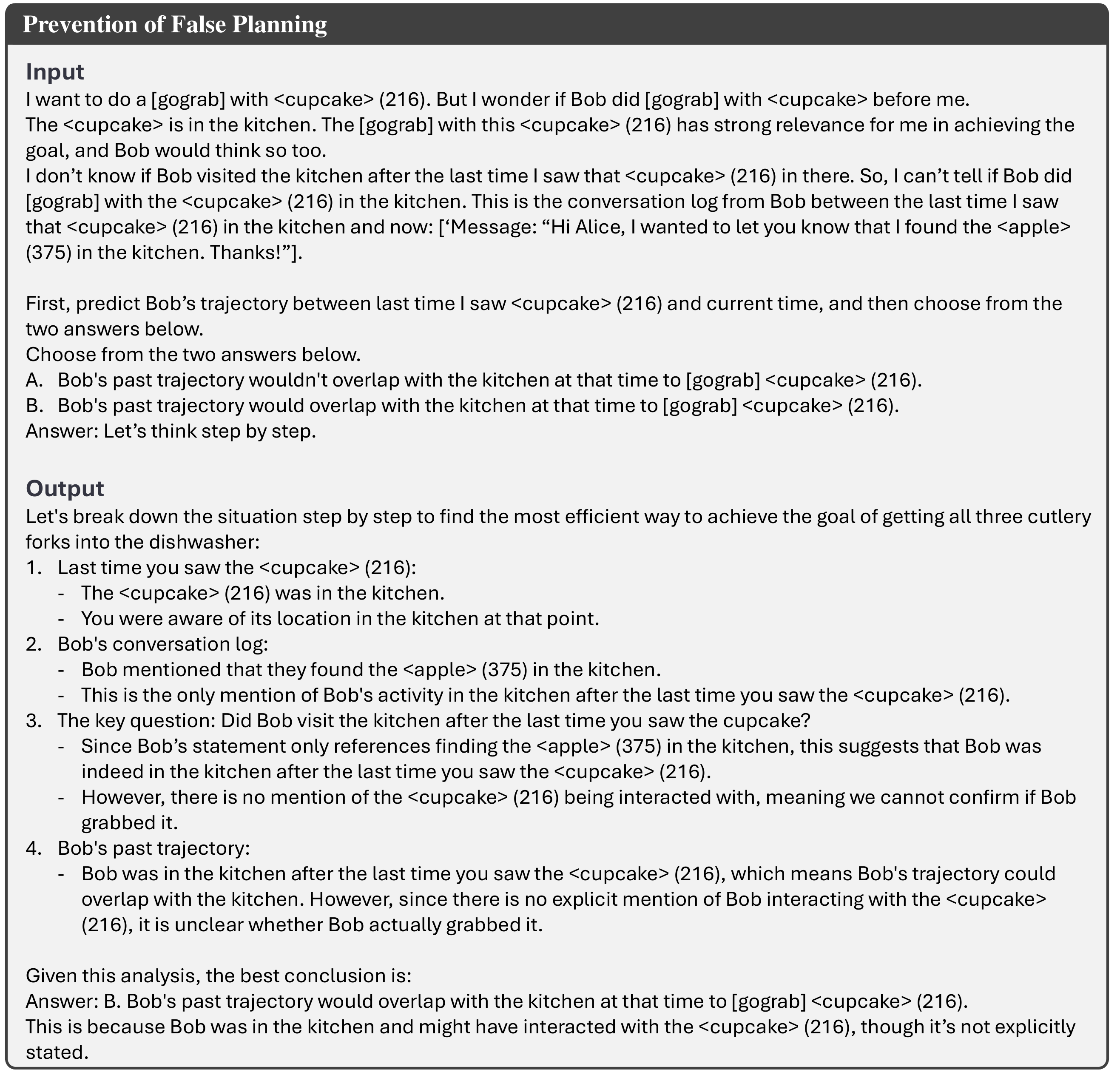}
    \caption{Prompt example for the prevention of false planning.}
\label{fig:prevention_of_false_planning}
\end{figure*}

In Figure~\ref{fig:prevention_of_false_planning}, Alice is about to grab the cupcake. Based on Bob's previous conversation, Alice knows Bob was searching for an apple in the kitchen. This implies Bob is in the kitchen where the cupcake is also located. Since the cupcake is an important object for both Alice and Bob in achieving their goals, Alice can infer that Bob may have already grabbed the cupcake. Using this information, Alice avoids executing [gograb] $<$cupcake$>$ (216) and instead performs Adaptive Planning to prevent false planning and generate a new optimal plan.

\onecolumn 

\onecolumn 
\newpage
\twocolumn 

\end{document}